%% file: acl.tex
\pdfoutput=1

\documentclass[11pt]{article}

\usepackage[table]{xcolor}
\input{math_commands.tex}

\usepackage{acl}

\usepackage{times}
\usepackage{latexsym}

\usepackage[T1]{fontenc}

\usepackage[utf8]{inputenc}

\usepackage{microtype}

\usepackage{multicol}
\usepackage{multirow}
\usepackage{pdfpages}

\usepackage{caption}
\usepackage{subcaption}
\usepackage{booktabs}

\definecolor{bold}{RGB}{154, 216, 245}
\definecolor{italic}{RGB}{201, 231, 245}
\definecolor{Gray}{gray}{0.9}

\usepackage{algorithm, algcompatible}
\usepackage{algpseudocode}
\usepackage{hyperref}
\usepackage{url}

\usepackage{booktabs} 
\usepackage{adjustbox}
\usepackage{graphicx}
\usepackage{subcaption}
\usepackage{multirow}
\usepackage{tabularx}
\usepackage{booktabs}
\usepackage{svg}
\usepackage{natbib}
\usepackage[normalem]{ulem}
\useunder{ine}{}{}
\usepackage{multicol}
\usepackage{pdfpages}

\usepackage{tcolorbox}

\tcbuselibrary{listings,breakable}

\tcbset{
  aibox/.style={
    width=\textwidth,
    top=0pt, bottom=0pt, left=5pt, right=5pt,
    colback=white,
    colframe=black,
    colbacktitle=black,
    center,
  }
}
\newtcolorbox{AIbox}[2][]{aibox,title=#2,#1}

\definecolor{aigold}{RGB}{244,210, 1} 
\definecolor{aigreen}{RGB}{210,244,211}

\definecolor{aired}{RGB}{255,180,181}

\newtcbox{\mybox}[1][green]{on line,
arc=0pt,outer arc=0pt,colback=#1!10!white,colframe=#1!50!black,
boxsep=0pt,left=0pt,right=0pt,top=0pt,bottom=0pt,
boxrule=0pt,bottomrule=0pt,toprule=0pt}

\usepackage{multirow}

\usepackage[utf8]{inputenc} 
\usepackage[T1]{fontenc}    
\usepackage{hyperref}       
\usepackage{url}            
\usepackage{booktabs}       
\usepackage{amsfonts}       
\usepackage{nicefrac}       
\usepackage{microtype}      
\usepackage{xcolor}         
\usepackage[normalem]{ulem}
\useunder{\uline}{\ul}{}
\usepackage{algorithm, algcompatible}
\usepackage{algpseudocode}
\usepackage{multicol}
\usepackage{multirow}
\usepackage{listings}

\usepackage{pdfpages}

\usepackage{caption}
\usepackage{subcaption}
\usepackage{booktabs}

\definecolor{bold}{RGB}{154, 216, 245}
\definecolor{italic}{RGB}{201, 231, 245}
\definecolor{Gray}{gray}{0.9}

\usepackage{algorithm, algcompatible}
\usepackage{algpseudocode}

\newcolumntype{L}{>{\centering\arraybackslash}m{3cm}}

\usepackage{multicol}
\usepackage{multirow}
\usepackage{pdfpages}

\usepackage{caption}
\usepackage{subcaption}
\usepackage{booktabs}

\title{Diffusion Language Models Generation Can Be Halted Early}

\author{Sofia Maria Lo Cicero Vaina$^*$, Nikita Balagansky$^*$, Daniil Gavrilov \\
    Tinkoff \\
  \texttt{s.lochichero@tinkoff.ru, n.n.balaganskiy@tinkoff.ai}, \texttt{d.gavrilov@tinkoff.ai}}

\begin{document}

\maketitle
\def\thefootnote{*}\footnotetext{These authors contributed equally to this work}\def\thefootnote{\arabic{footnote}}

\begin{abstract}
Diffusion Language models (DLMs) are a promising avenue for text generation due to their practical properties on tractable controllable generation. They also have the advantage of not having to predict text autoregressively. However, despite these notable features, DLMs have not yet reached the performance levels of their autoregressive counterparts. One of the ways to reduce the performance gap between these two types of language models is to speed up the generation of DLMs. Therefore, we propose a novel methodology to address this issue in this work. It enables the execution of more generation steps within a given time frame, leading to higher-quality outputs. Specifically, our methods estimate DLMs completeness of text generation and allow adaptive halting of the generation process. We evaluate our methods on Plaid, SSD, and CDCD DLMs and create a cohesive perspective on their generation workflows. Finally, we confirm that our methods allow halting these models and decrease the generation time by $10$-$40$\% without a drop in the quality of model samples.
\end{abstract}

\section{Introduction}

Exploring Large Language Models (LMs) is a dominant research direction in NLP. The two primary methods of training LMs for NLP are autoregressive training \citep{gpt, t5, palm} and masked language modeling \citep{bert, deberta, roberta, albert}. 

The exploration of alternative models, such as diffusion models \citep{ddpm, ddim}, is a promising avenue for research as diffusion allows native non-causal conditioning and simplified controllable generation methods \citep{glide}. In recent works, with models such as "Diffusion LM" and Plaid \citep{diffusion_lm, plaid}, Simplex-based Diffusion Language Model (SSD) \citep{ssd}, GENIE \citep{genie}, and Continuous Diffusion for Categorical Data (CDCD) \citep{continuous_diffusion} being introduced, indicating an emerging interest for using diffusion models in text generation. 

A crucial distinction between autoregressive LMs and diffusion language models (DLMs) lies in their modeling approaches. Autoregressive LMs predominantly adhere to the common probabilistic model. In contrast, DLMs exhibit substantial divergence in their application for modeling categorical data. When exploring DLMs, it is essential to consider the lack of connectivity between such models. The majority of comparisons between them have primarily focused on evaluating sample quality \citep{plaid, ssd}. 

While it is essential to study the sample quality of DLMs, it does not further our understanding of the differences between these DLMs models. This work addresses this issue and evaluates various DLMs with a unified view of their generation process. Given this unified view, we study the dynamics of the generation process within different DLMs and focus on the changes in the samples during that process. Given this dynamic perspective, we observed that performing dynamic halting of the generation process with various DLMs is possible.

The main contributions of this paper can be summarized as follows:
\begin{itemize}
\item We showed that the generation process of most DLMs for general text generation can be halted, which makes it possible to implement an early, faster sample generation without compromising quality.
\item To the best of our knowledge, we were the first to evaluate DLMs with adaptive Early Exiting \citep{act}. In this paper, we introduced three adaptive criteria inspired by the ones used for text classification \citep{fastbert, pabee, fpabee}. 
\item We evaluated these criteria and provided empirical evidence of their efficiency. More concretely, we observed that performing an early exit did not hurt the quality of DLM samples, thus allowing faster generation.
\end{itemize}

\section{Related Work}
\label{section:related}

\subsection{Diffusion Language Models}
Diffusion models have shown promise in discrete data tasks such as image generation and NLP. However, most of the recent works were not evaluated with unconditional text generation \citep{analog_bits, sundae, diffuser, diffuseq, seqdiffuseq, genie}. 

The "Diffusion LM" by \citet{diffusion_lm} was a stride towards a generalized LM for unconditional sampling but was limited by not being trained on large datasets. Models like Self-conditioned Embedding Diffusion (SED) and Continuous Diffusion for Categorical Data (CDCD) did employ extensive pre-training \citep{sed, continuous_diffusion}, but without releasing weights, necessitating training from scratch for any comparison. 

In contrast, the openly available and pre-trained Simplex-based Diffusion Language Model (SSD) and Plaid models present a practical advantage for comparative studies \citep{ssd, plaid}. These diffusion models are appealing to use for comparison due to the different approaches used for training them. For instance, while CDCD utilizes a score interpolation objective, SSD works with a simplex-based method. On the other hand, Plaid is defined with a Variational Lower Bound objective \citep{vae, variational_diffusion}.

\subsection{Early Exiting Methods}

The early exit technique is an approach for reducing computational load \citep{act}. Having a sequential process (for example, a model with several layers or a model that is evaluated several times), early exiting approaches aim to reduce the number of steps in this process. 

In the case of multi-layered modes such as Transformers, one could reduce the number of evaluated layers since intermediate hidden states maintain consistent shapes across layers. As a result, early exiting has become a standard technique for downstream tasks with pre-trained LMs \citep{pabee, fastbert, palbert, fpabee}.

When speaking of diffusion models (including language models), it is possible to halt the generation process since it consists of multiple forward passes of a model. It was previously explored with image generation \citep{es-ddpm}. However, early exiting was not studied with DLMs. Due to the specific nature of parametrization and definition of a probabilistic model with DLMs, it is especially interesting to understand the possibility of performing early exiting with them.

\section{Preliminaries}
\subsection{Diffusion Language Models}

DLMs have emerged as an innovative approach model text within NLP and deep learning. While these models might be less familiar to some practitioners accustomed to traditional architectures like RNNs \citep{lstm} or even the more recent Transformer models, they offer a unique perspective on text generation. We will delve into the operational principles of DLMs, focusing on three specific methodologies: CDCD, Plaid, and SSD.

\subsubsection{Token Representation and Embeddings}

In DLMs, we commence with token sequences that are standard in NLP. Tokens $\vx$ are projected into continuous vector space through an embedding matrix, similar to the initial layers of a Transformer model. We denote these initial, noise-free embeddings as $\mX_0$. Reflecting their nature in the context of diffusion models, these tokens subsequently undergo a noising process, resulting in time-dependent embeddings $\mX(t)$, with the noise level modulated by the timestep $t$.

\subsubsection{Continuous Diffusion for Categorical Data (CDCD)}

CDCD introduces a novel take on denoising where noisy token embeddings are progressively refined to approximate the original, clean token embeddings. At each diffusion step, characterized by time $t$, the model estimates the denoised embeddings $\hat{\mX}_0(t)$ via calculation of a probability distribution over the tokens $p(\vx|\mX(t), t)$. This distribution is transformed from logits via a softmax function, typical of categorical outputs in NLP models, and used to compute a cross-entropy loss, serving as feedback to the model for its predictions.

\subsubsection{Plaid}

Plaid adopts a different mechanism based on the Variational Lower Bound (VLB) \citep{vae}. By optimizing the VLB, Plaid estimates the denoised token embeddings and defines a forward process in which noise is added to the embeddings. The model is then guided to reverse this process to recover the denoised tokens. This concept resembles VAEs, where the latent space is sampled and decoded back to the input space, though with specific nuances and formulations suitable for diffusion processes.

While the resulting loss differs from CDCD, Plaid also models the categorical distribution over possible tokens $p(\vx|\mX(t), t)$ at each generation step, making it possible to study the dynamics of its properties.

\subsubsection{Simplex-Based Diffusion Language Model (SSD)}

SSD uses almost-one-hot encodings to represent tokens. For token $\vx_i$, its continuous representation is defined as $\mX_{i,j} = K$ if $\vx_i = \mV_j$, and $\mX_{i,j}= -K$ otherwise. $K \in \R^+$ here is a hyperparameter, and $\mV$ is a vocabulary.

In the diffusion process specific to SSD, noise is incrementally added to these representations, blurring their sharp distinctions over time until they resemble a normal distribution in the logit space. .

The training of the SSD model involves optimizing a loss that evaluates the probabilities of predicting the correct next token given the context of preceding tokens and the current noisy state of token embeddings. The loss encourages the model to accurately predict the token distribution even as the representations become increasingly noisy, a challenge akin to predicting the next word in a sentence as the context becomes less defined.

\section{Early Exiting with DLMs}
\label{section:method}
While CDCD, Plaid, and SSD define different views on training DLMs, they share a similarity in how distribution on denoised text is defined. More concretely, they all define a categorical distribution over possible embeddings (and thus over possible tokens) $p(\vx|\mX(t), t)$. This fact leads to the question of \textit{how the distribution of possible tokens changes with time}. 

A natural way to assess the dynamics of the generation process is to think of it in terms of Adaptive Computation Time \citep{act} (i.e., early exiting methods). The concept of early exiting is a well-established practice in various research fields of Deep Learning \citep{act, fastbert, pabee, palbert, act}. Consequently, there are numerous methods available for performing an early exit. 

\textbf{Entropy criterion}, described by \citet{fastbert}, is one of the most common early exit techniques. This method performs an exit when entropy drops below a certain threshold. A major downside of the entropy criterion is that it disregards the output dynamics, resulting in overly confident classifiers.

\textbf{Patience-based criterion}, as proposed by \citep{pabee}, addresses the limitations of the Entropy criterion. It is formulated as follows: if the classifier predictions remain unchanged for a series of $t$ consecutive steps, the model initiates an exit. A notable drawback of Patience is its insensitivity to the scale of the changes in underlying distribution. A small change in distribution could lead to a change of prediction, thus not allowing a method to halt, and vice versa. Another drawback of this approach is that it requires a substantial number of steps for the patience value to become meaningful, which is not ideal when the goal is to minimize the number of steps.

\textbf{KL criterion} overcomes the drawbacks of the Patience-based criterion \citep{fpabee}. This criterion triggers an exit when the KL Divergence between the current diffusion step's distribution and the previous one falls below a certain threshold. 

\textbf{Fixed step criterion} is a simple procedure of exiting after a fixed computational step without conditioning on any statistics of the evaluated model. Though this criterion does not include any adaptivity depending on the model statistics, we conducted experiments with it for the complete picture.

When speaking of DLMs, Entropy and KL criteria could be applied as is (see Appendix Algorithms \ref{alg:e}, \ref{alg:kl}). For the Entropy criterion, we assess the entropy of $p(\vx|\mX(t), t)$ during the generation and halt computation once it falls below the pre-determined threshold. For the KL criterion, we evaluate KL Divergence between $p(\vx|\mX(t), t)$ in two sequential steps. As for the Entropy criterion, we halt generation once the KL Divergence value becomes smaller than the threshold value. 

Recent works utilized patience criterion based on change of predictions made by model \citep{pabee}. To use it with DLMs, we assessed the measure of changed tokens after a generation step (namely \textit{token swithces}). A small number of altered tokens could indicate that the generation process converges. Once the number of altered tokens falls below the threshold value for a sequence of generation steps, we halt generation (further details are provided in Appendix Algorithm \ref{alg:p}). 

Finally, we also evaluate a fixed step criterion for which we halt the generation regardless of the statistics of the model. While this criterion could be seen as overly simplified, it is still helpful to understand how the generation process evolves with time.

In the following sections, we first study the ability to apply these criteria to DLMs and then evaluate their performance.

\section{Experiments}
\label{section:early_exit}

\subsection{Experimental Setup}

Performance evaluations were carried out using Autoregressive Negative Log-Likelihood (AR-NLL) \citep{continuous_diffusion} with GPT-Neo-1.3B \citep{gpt-neo} as a third-party language model to compute log-likelihood for generated samples. Additionally, diversity was measured through the count of distinct N-grams across five samples from a single prompt, denoted as $\text{dist}_{N}$, and Self-BLUE \citep{sblue}. For some experiments, we also assessed MAUVE metric \citep{mauve} for measuring the quality of generated texts and Zipf's coefficient to evaluate generated token statistics. Collectively, these metrics provide a comprehensive assessment of the text-generation capabilities of the DLMs.

Since the original tools for CDCD were not shared publicly, we developed our rendition, the Democratized Diffusion Language Model (DDLM). For the details on how we replicated and trained DDLM, refer to the Appendix Section \ref{section:ddlm-training}. In the following sections, we will use DDLM to understand the generation process of the CDCD framework.

All models underwent testing on the validation set from the C4 dataset \citep{t5}. Evaluations were performed under unconditional settings and with a 32-token prefix prompt, defaulting to the unconditional setup unless otherwise specified.

Note that there is no standard for the initial number of generation steps for text generation with DLMs \citep{continuous_diffusion, plaid, ssd}. To navigate this, we used $200$ generation steps for experiments to understand the statistical characteristics of the models' hidden states. For quality assessment of the generated text, we decided on $1000$ steps. The outcomes of these evaluations, comparing different models at varying step counts, can be found in the Appendix Table \ref{tab:final}.

\begin{figure}[h!]
  \centering
  
    \medskip
        \begin{subfigure}[t]{.9\linewidth}
    \centering\includegraphics[width=\linewidth]{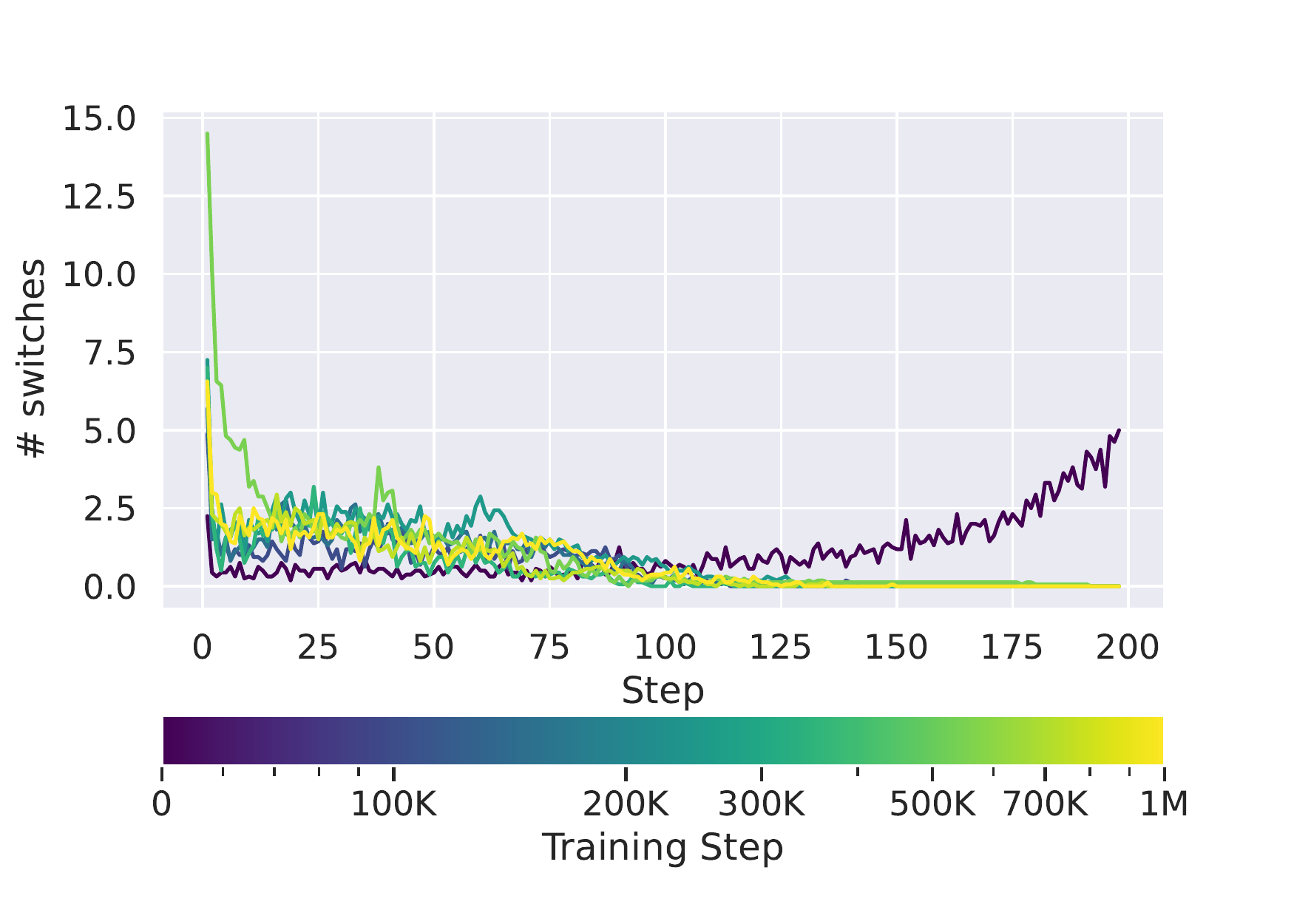}
    \caption{}
  \end{subfigure}
      \begin{subfigure}[t]{.9\linewidth}
    \centering\includegraphics[width=\linewidth]{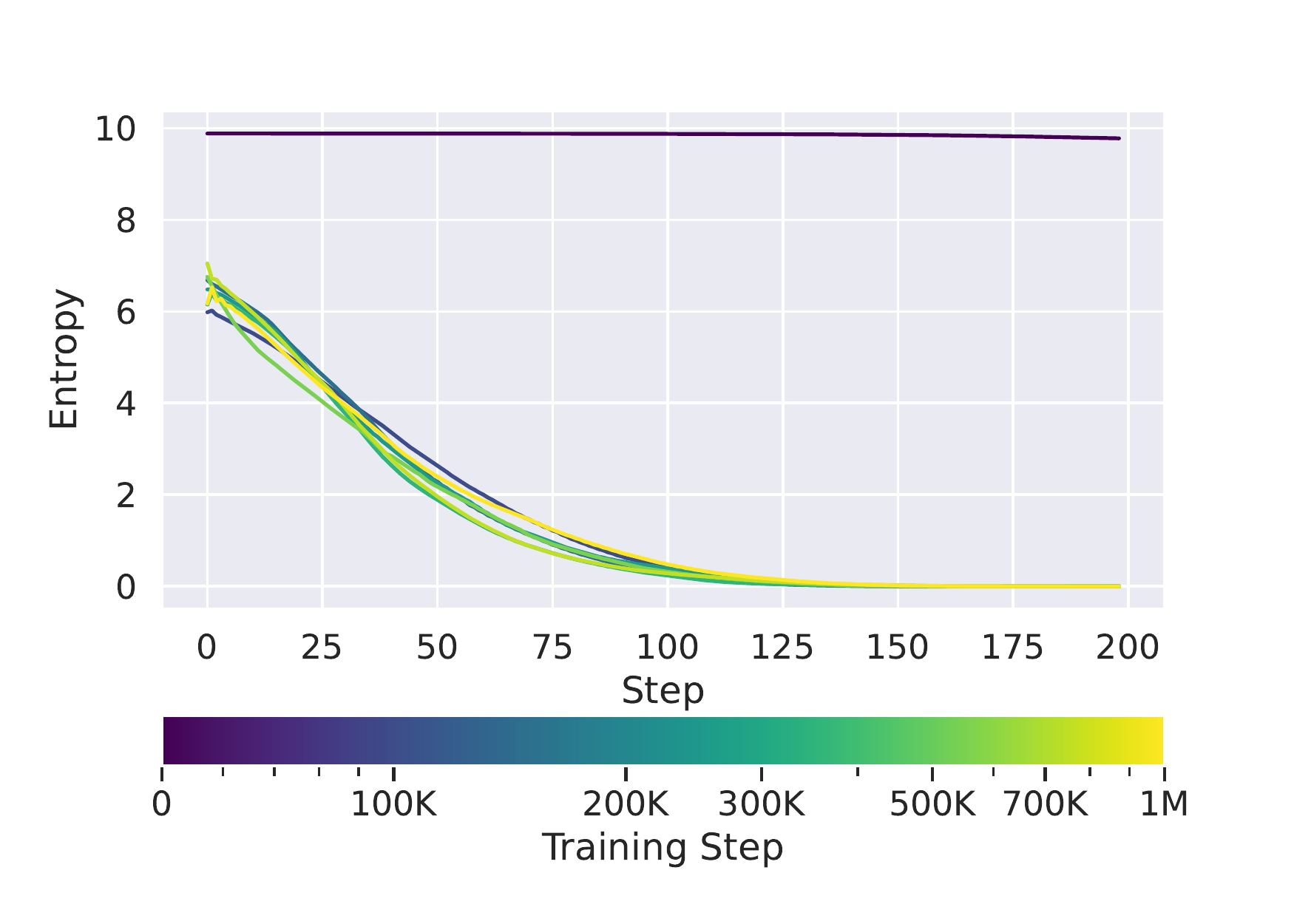}
    \caption{}
  \end{subfigure}

  \caption{(a) The number of token switches and (b) the entropy of $p(\vx|\mX(t), t)$. Color represents the training step, while the x-axis is the diffusion generation step. The trained model reaches the minimum entropy value before the generation process ends, and the resulting samples remain unchanged. This result indicates the possibility of performing an Early Exit from DLM generation without losing the quality of samples. See Section \ref{section:emerging} for more details.}
  \label{fig:switches}
\end{figure}

\begin{figure*}[h!]
  \centering
  
    \medskip
        \begin{subfigure}[t]{.49\linewidth}
    \centering\includegraphics[width=\linewidth]{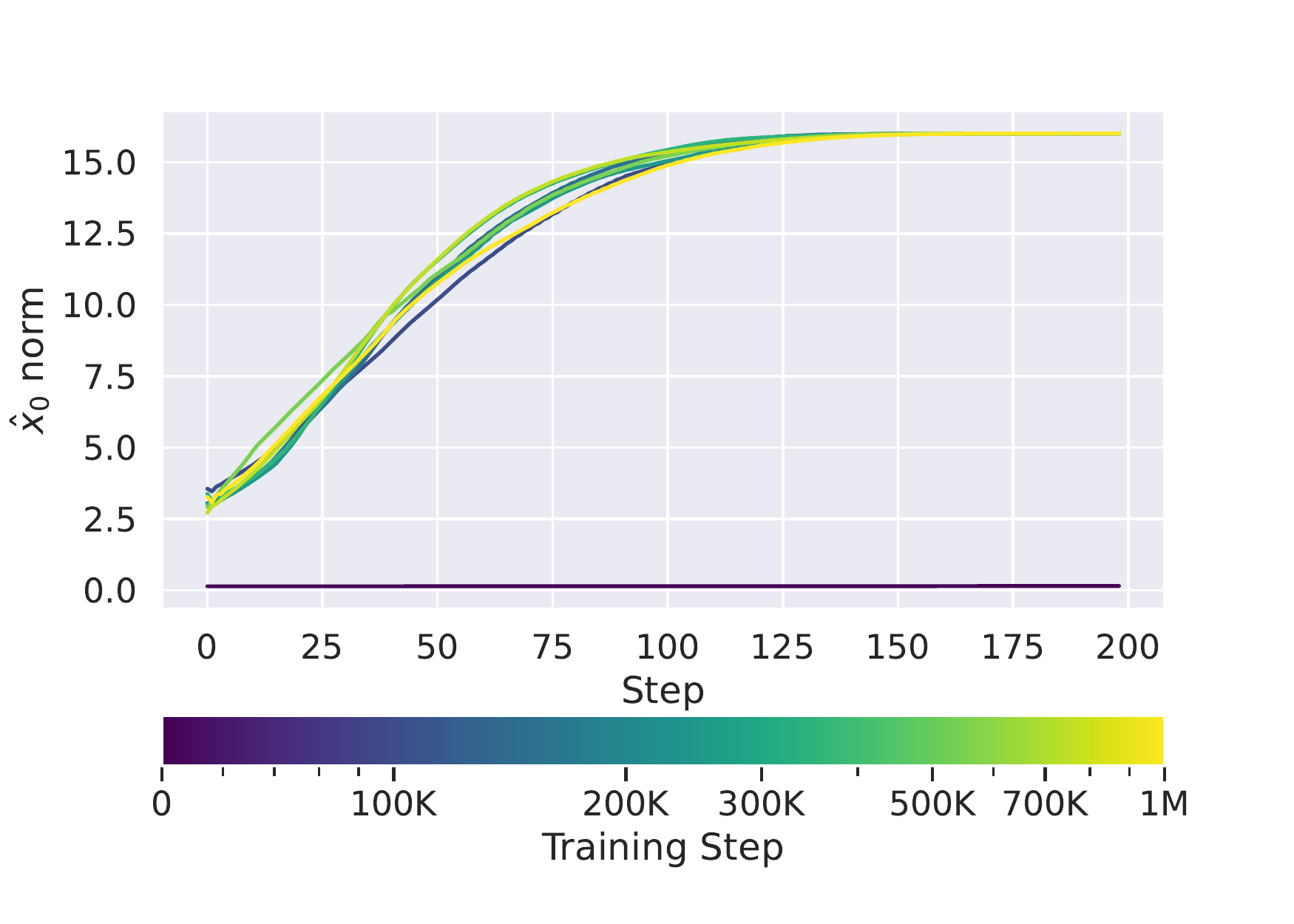}
    \caption{}
  \end{subfigure}
      \begin{subfigure}[t]{.49\linewidth}
    \centering\includegraphics[width=\linewidth]{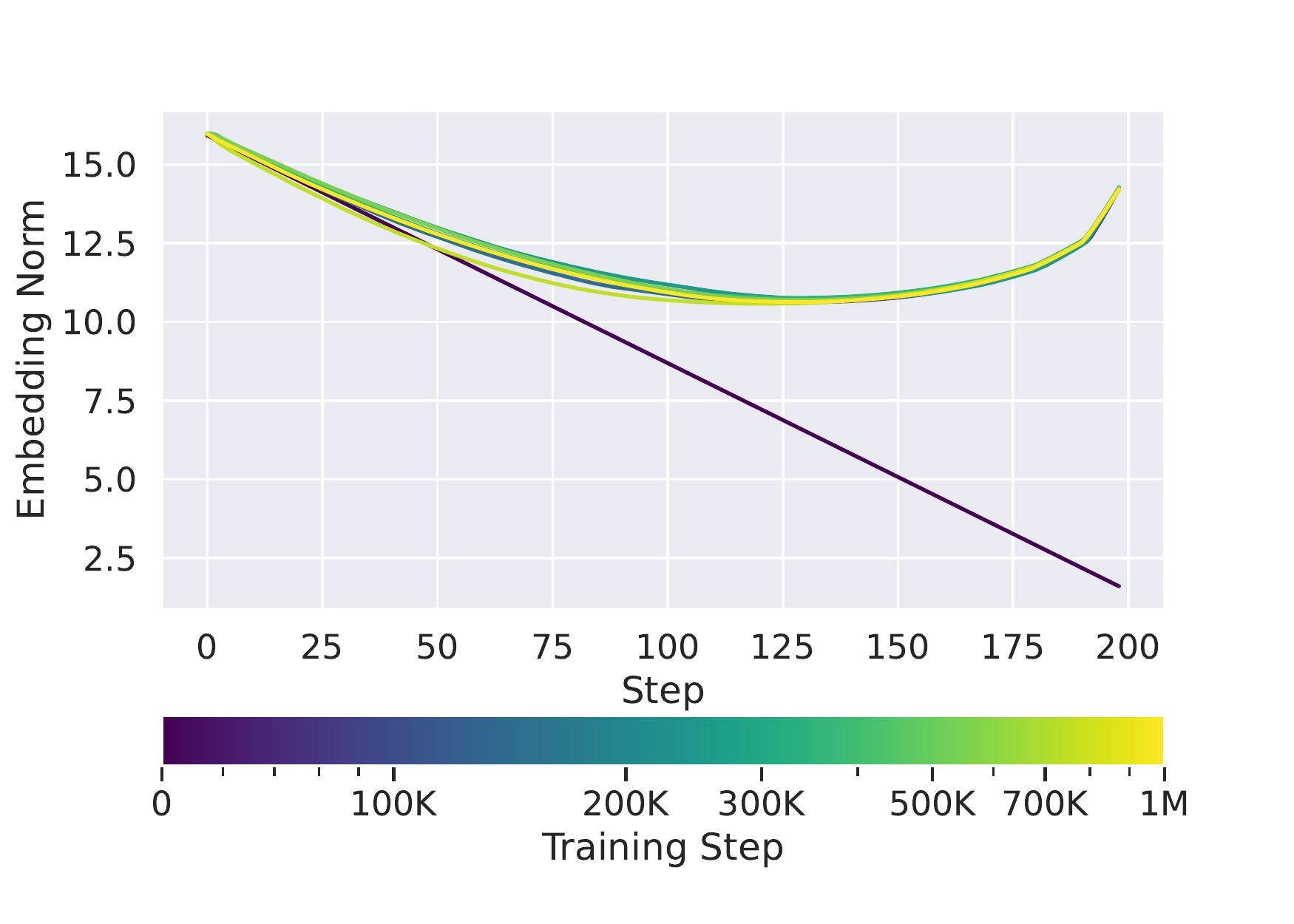}
    \caption{}
  \end{subfigure}
        \begin{subfigure}[t]{.49\linewidth}
    \centering\includegraphics[width=\linewidth]{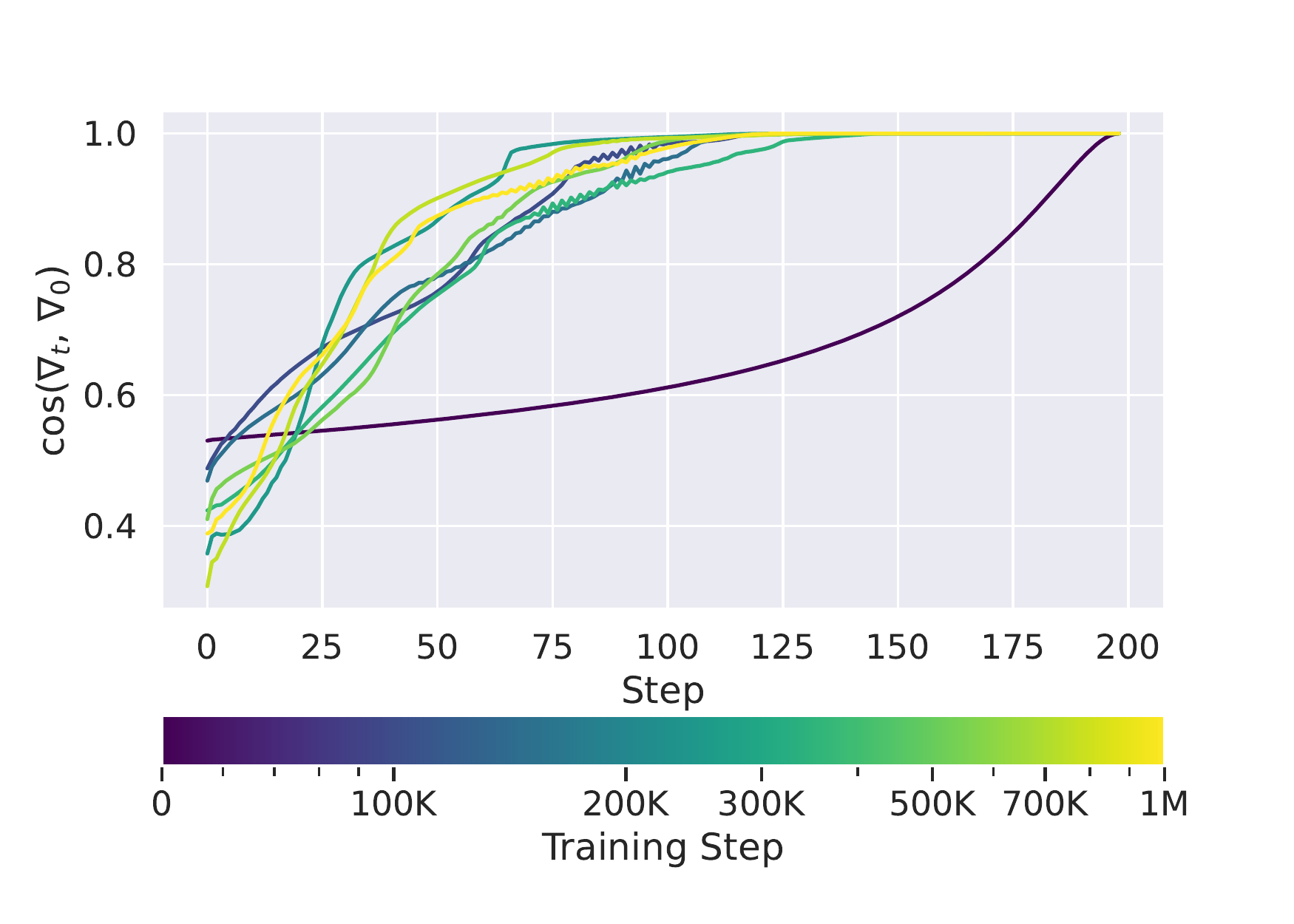}
    \caption{}
  \end{subfigure}
        \begin{subfigure}[t]{.49\linewidth}
    \centering\includegraphics[width=\linewidth]{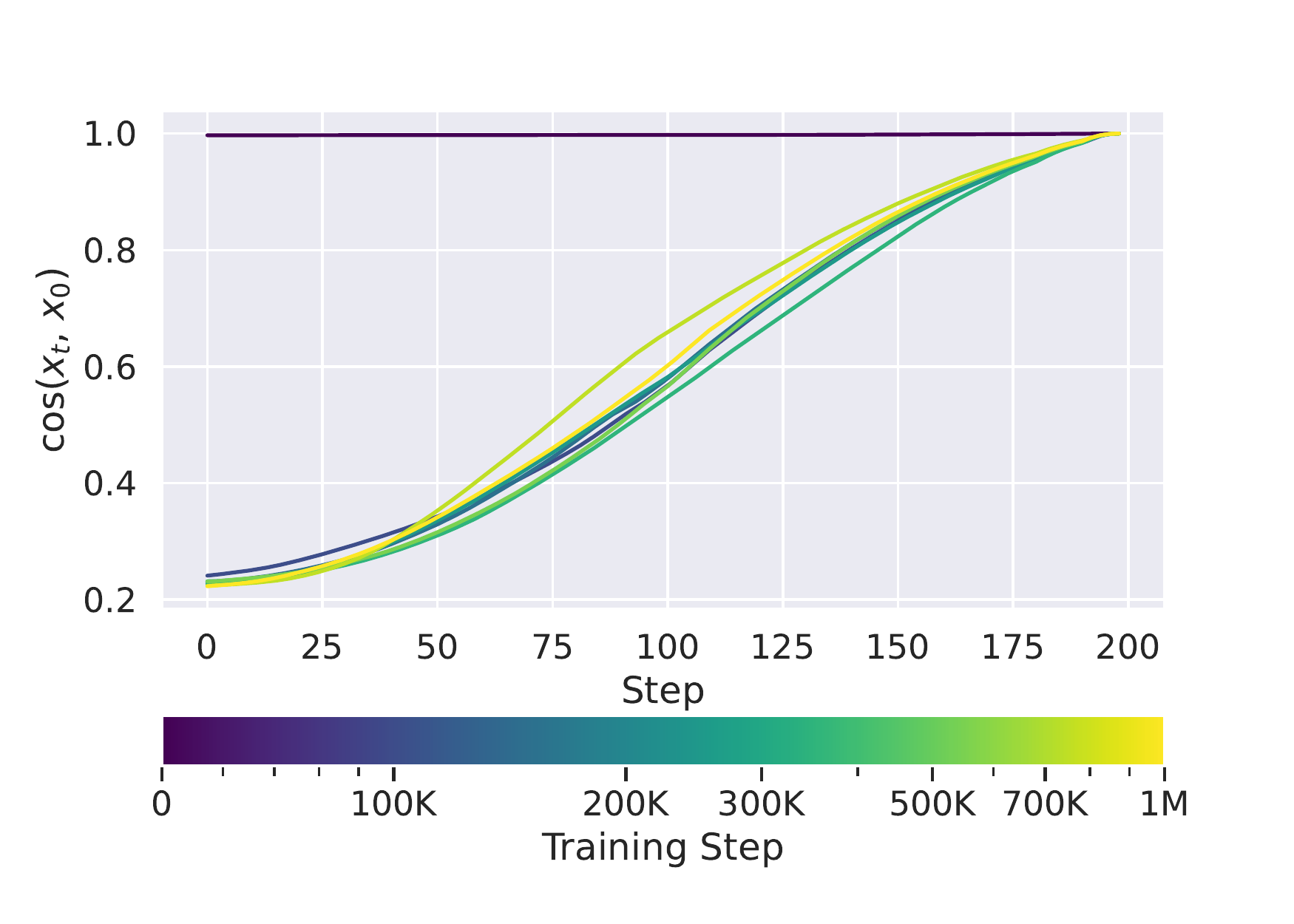}
    \caption{}
  \end{subfigure}
    \caption{(a) The L2 norm of embeddings $||\hat{\mX}_0||_2(t)$, (b) the L2 norm of embeddings $||\mX||_2(t)$, (c) $\cos$ of the angle between score estimation $\hat{\mS}$ and final score in the end of generation, and (d) $\cos$ of the angle between embedding $x$ and final embedding in the end of generation. Color represents the training step, while the x-axis is the diffusion generation step. Beyond step 100, the change in scoring angle ceases, suggesting the model has determined the optimal direction for enhancing the embedding midway through generation. See Section \ref{section:emerging} for more details. }
    \label{appendix-fig:cos-base}
\end{figure*}

\subsection{Emergence of Early Exiting Behavior}
\label{section:emerging}
To explore token behavior during generation, we analyze the number of token switches in DDLM. We evaluate token switches at different pre-training checkpoints and at each time step $t$ during generation for DDLM. Additionally, we examine the entropy of the embedding prediction $p(\vx|\mX(t), t)$. Sequences with $200$ steps are sampled for this analysis (see Figure \ref{fig:switches}). Interestingly, the model shows zero token switches after approximately the 100th sampling step. This suggests a potential for adaptive early exiting in DDLM generation since, for nearly half of the generation steps, the sampling algorithm only made minor adjustments to predicted embeddings without changing the generated tokens. Depending on the sequence, adaptive early exiting will make it possible to dynamically evaluate when we can halt the generation process, potentially greatly reducing the computations needed for sampling.

\begin{figure}[h!]
  \centering
  
    \includegraphics[width=0.99\linewidth]{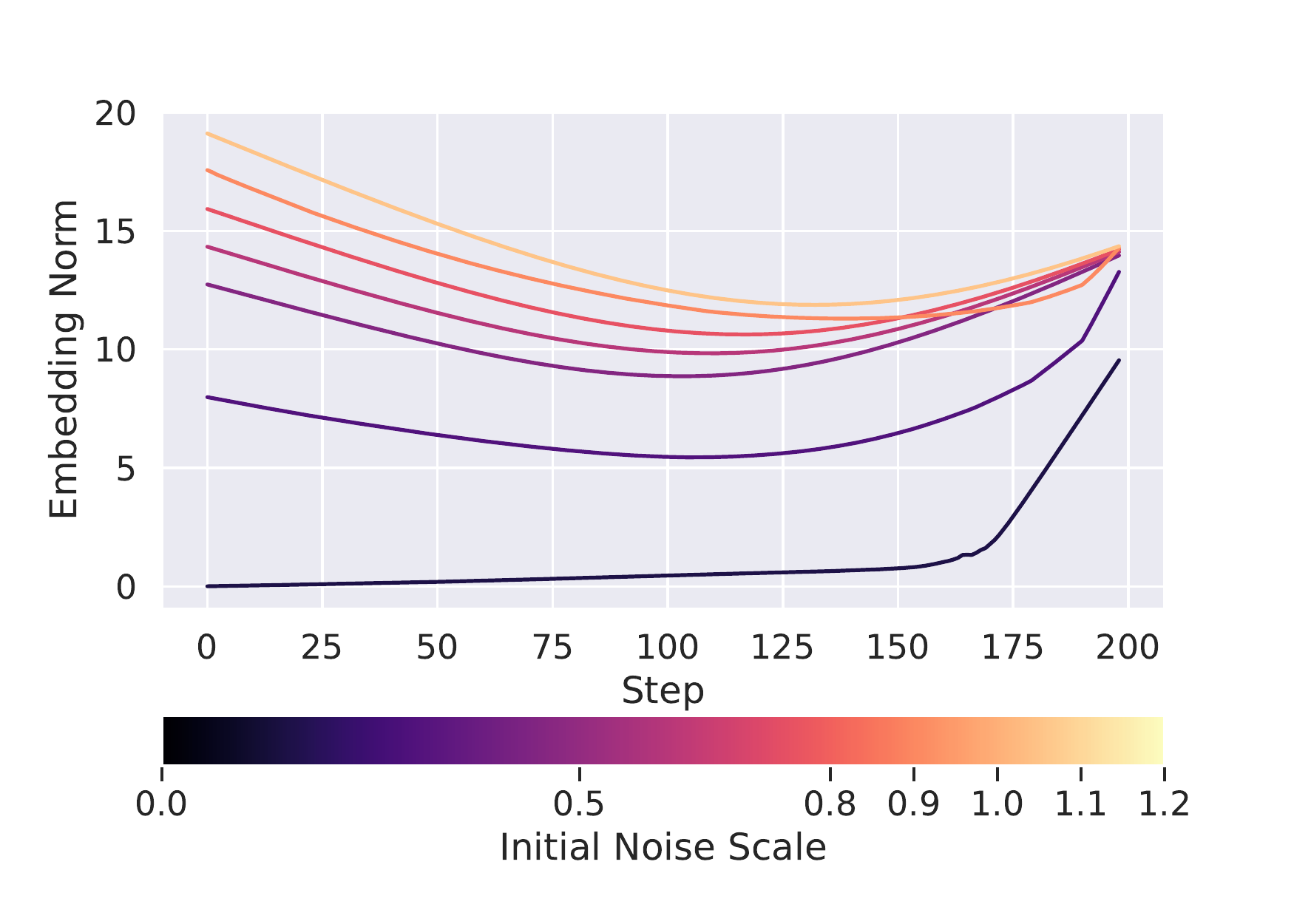}%
    \caption{The L2 norm of embeddings $||\mX||_2(t)$ during the generation process for different initial scales of $||\mX||_2$ for DDLM. Color represents the initial noise scale, while the x-axis is the diffusion generation step. A lower initial noise scale allows us to reach a minimum of the $||\mX||_2$ L2 norm faster, indicating the dependence of Early Exiting performance with the initial noise scale. See Section \ref{section:early_exit} for more details.}
    \label{fig:scale}
\end{figure}


\begin{figure*}

    \medskip
        \begin{subfigure}[t]{.32\linewidth}
    \centering\includegraphics[width=\linewidth]{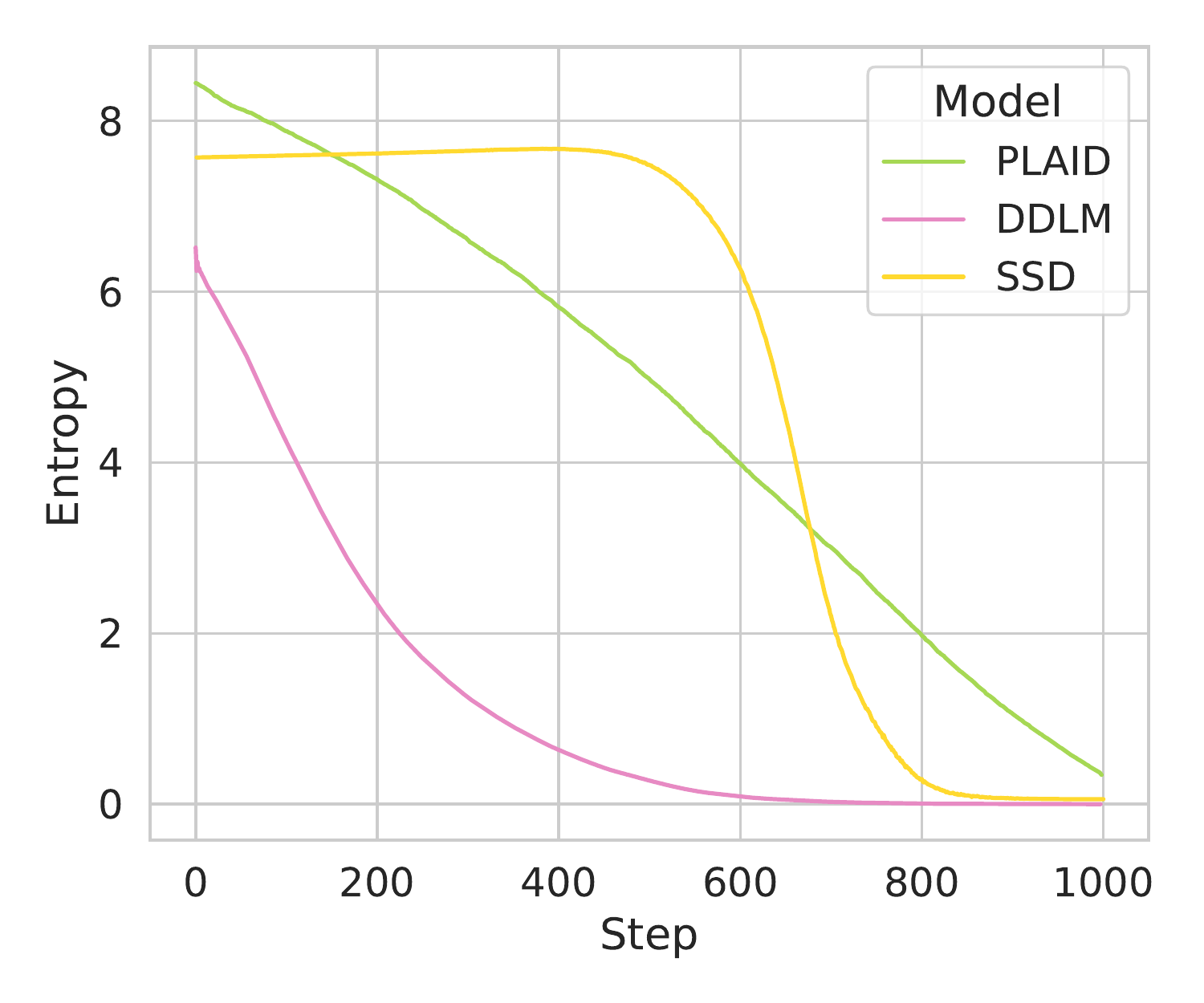}
    \caption{}
  \end{subfigure}
      \medskip
        \begin{subfigure}[t]{.32\linewidth}
    \centering\includegraphics[width=\linewidth]{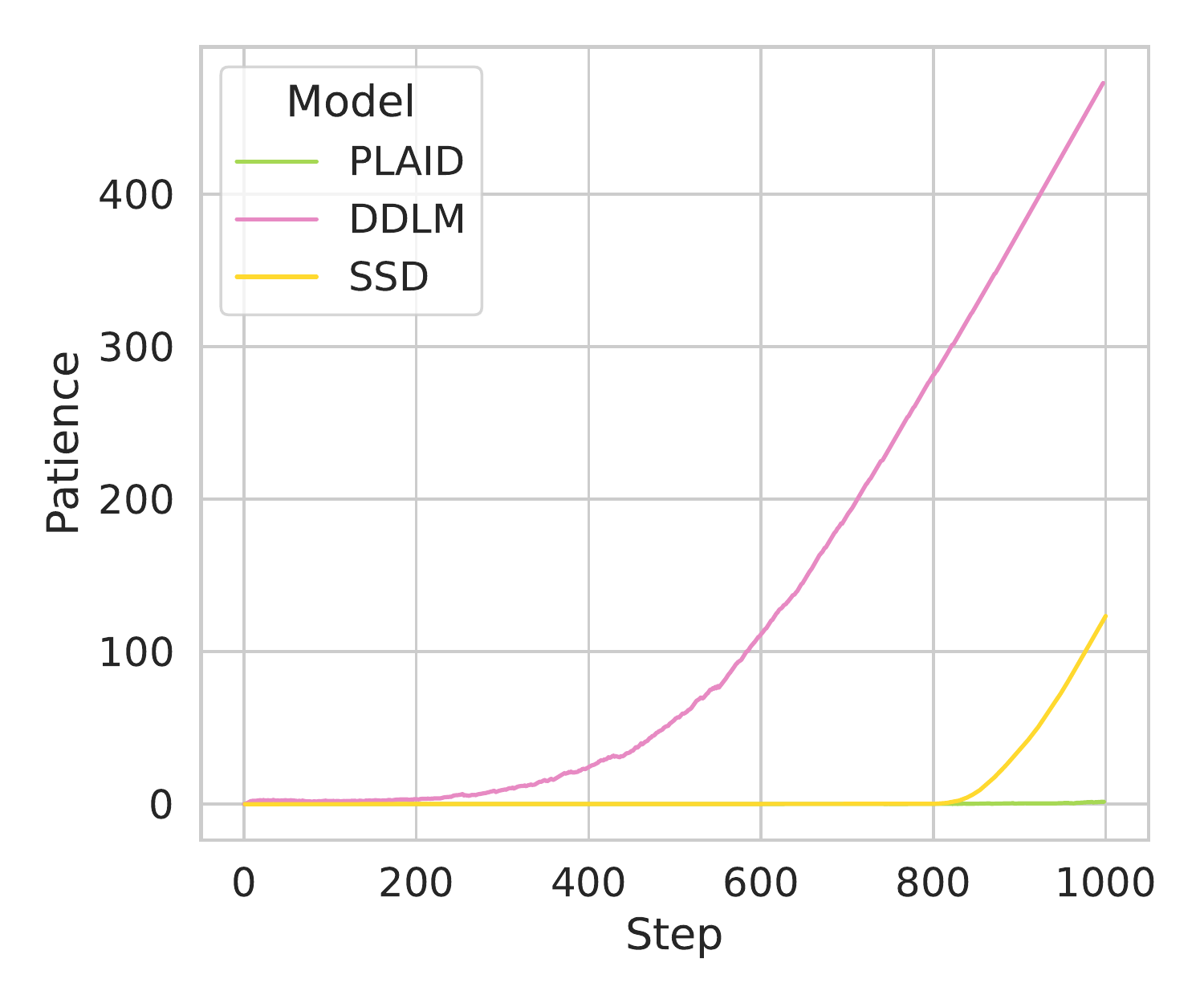}
    \caption{}
  \end{subfigure}
      \medskip
        \begin{subfigure}[t]{.32\linewidth}
    \centering\includegraphics[width=\linewidth]{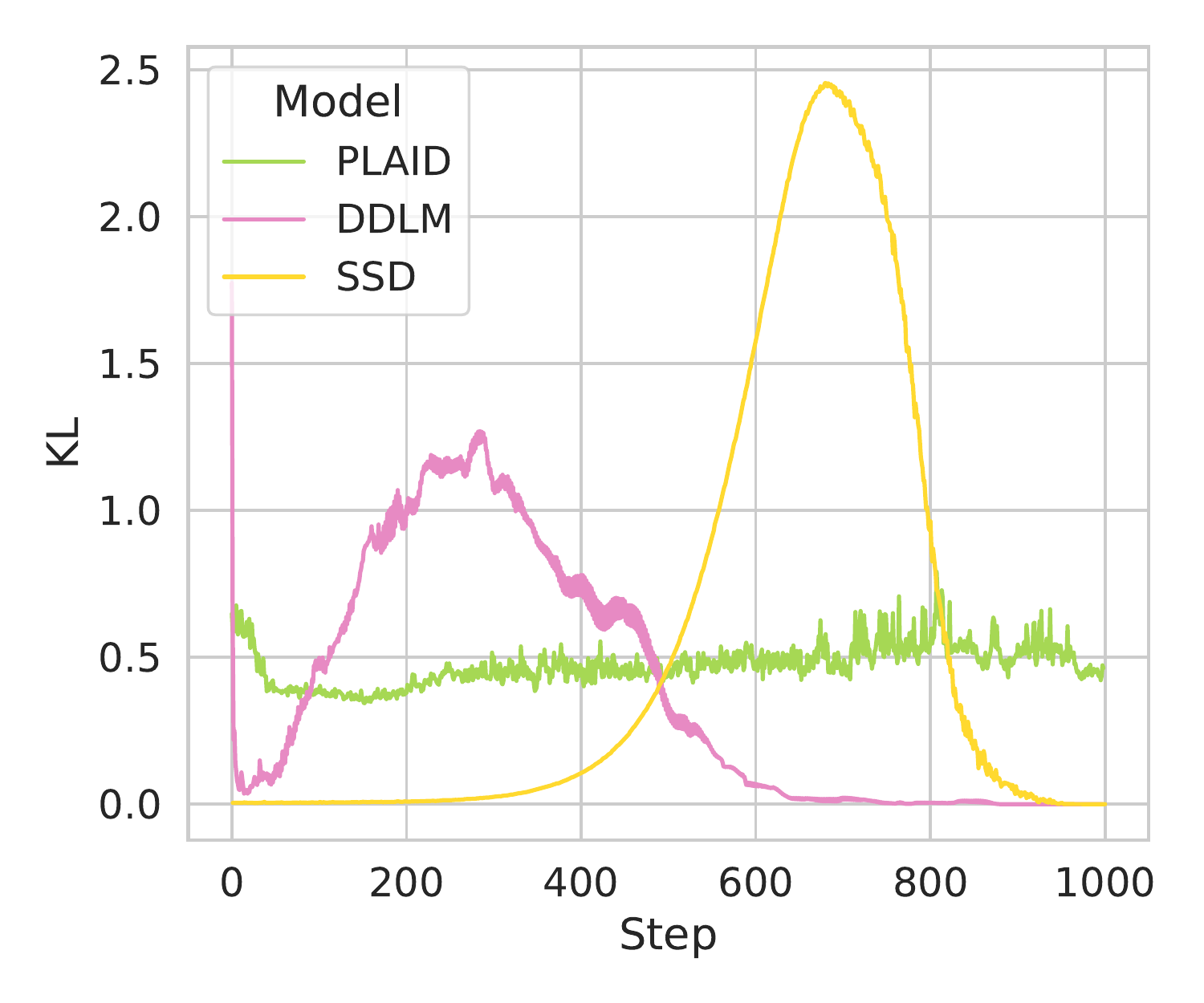}
    \caption{}
  \end{subfigure}

    \caption{(a) Entropy, (b) unchanged step count, and (c) KL-Divergence are used for different criteria in DDLM, SSD, and Plaid. Generation is halted when the threshold values are met. DDLM reaches the threshold early on, while SSD does so later. The result indicates that DDLM could allow early stopping in text generation. SSD reaches a stopping point after 800 steps of the total 1000. In contrast, Plaid's entropy decreases steadily, and other measures stay the same, hinting that it might not perform well with adaptive early stopping techniques (though still capable of performing a fixed step halting). See Section \ref{section:method} for more details.}
    \label{fig:mean-criteria}
\end{figure*}

To understand why the trained model tends towards minimal token switches early on in the generation process, we examined the L2 norm of $\hat{\mX}_0(t)$ and $\mX(t)$ during generation\footnote{For the reader's convenience, it is essential to remember that $\mX(t)$ are embeddings passed to the model as an input. At the same time, $\hat{\mX}_0(t)$ are embeddings produced by the model to estimate the score function.} (refer to Figure \ref{appendix-fig:cos-base}). We found that $\hat{\mX}_0$ rapidly reaches an L2 norm of $16$, the L2 norm of normalized embeddings during pre-training. This aligns with our observation of the entropy of $p(\vx|\mX(t), t)$ reaching near-zero values within $100$ generation steps. Fascinatingly, the L2 norm of $\mX$ first reduces and then increases from its large initialization value, suggesting that $\mX$ travels from one point on the embedding sphere surface to another via its interior.

\begin{table}[ht!]
\begin{center}
\begin{small}

\begin{tabular}{r|ccccc}
\toprule
Noise & AR-NLL & $\text{dist}_1$ & $\text{dist}_2$ & $\text{dist}_3$ & s.-BLEU \\
\midrule
0.0   & 0.44   & 0.00            & 0.00            & 0.00            & 1.00      \\
0.5   & 3.10   & 0.24            & 0.47            & 0.60            & 0.58      \\
0.8   & 3.50   & 0.41            & 0.74            & 0.84            & 0.47      \\
0.9   & 3.62   & 0.48            & 0.83            & 0.92            & 0.49      \\
1.0   & 3.72   & 0.49            & 0.86            & 0.94            & 0.48      \\
1.1   & 3.86   & 0.51            & 0.88            & 0.90            & 0.47      \\
1.2   & 4.01   & 0.52            & 0.89            & 0.95            & 0.44    \\
\bottomrule
\end{tabular}
  \caption{Performance of DDLM depending on the initial noise scale of $\mX$. Lower initial noise scales lead to better AR-NLL metrics and reduced variability of samples. See Section \ref{section:early_exit} for more details.}%
    \label{table:ranges}
\end{small}
\end{center}
\end{table}

To support this hypothesis, we evaluate the $\cos$ between score $\hat{\mS}(\mX(t), t) = \frac{ \hat{\mX}_0(\mX(t), t) - \mX(t)}{t^2}$ with final score $\hat{\mS}_0$ \citep{karras}, and the $\cos$ between $\mX$ with final $\mX_0$ during the generation process. After the 100th step, the scoring angle stops changing, indicating that the model settles on the final embedding improvement direction of mid-generation. This constant direction forces $\mX$ to the embedding sphere boundary, leading to high-confidence results and near-zero token switches.

Empirical evidence suggests that $\mX$ traverses between two points on the surface of a sphere via its interior. By reducing the initial noise scale, we can adjust the trajectory of $\mX$. See Figure \ref{fig:scale} and Table \ref{table:ranges} for our results. We find that a lower initial noise scale makes it possible for $||\mX||_2$ to reach its minimum value during generation more quickly. However, this approach limits the variability of samples. While our findings show that using a noise scale of $0.9$ is optimal, we will use a scale of $1.0$ in later experiments for convenience.

\subsection{Exploring Early Exit Criteria}
\label{section:method}

\begin{figure*}[ht!]

    \medskip
        \begin{subfigure}[t]{.32\linewidth}
    \centering\includegraphics[width=\linewidth]{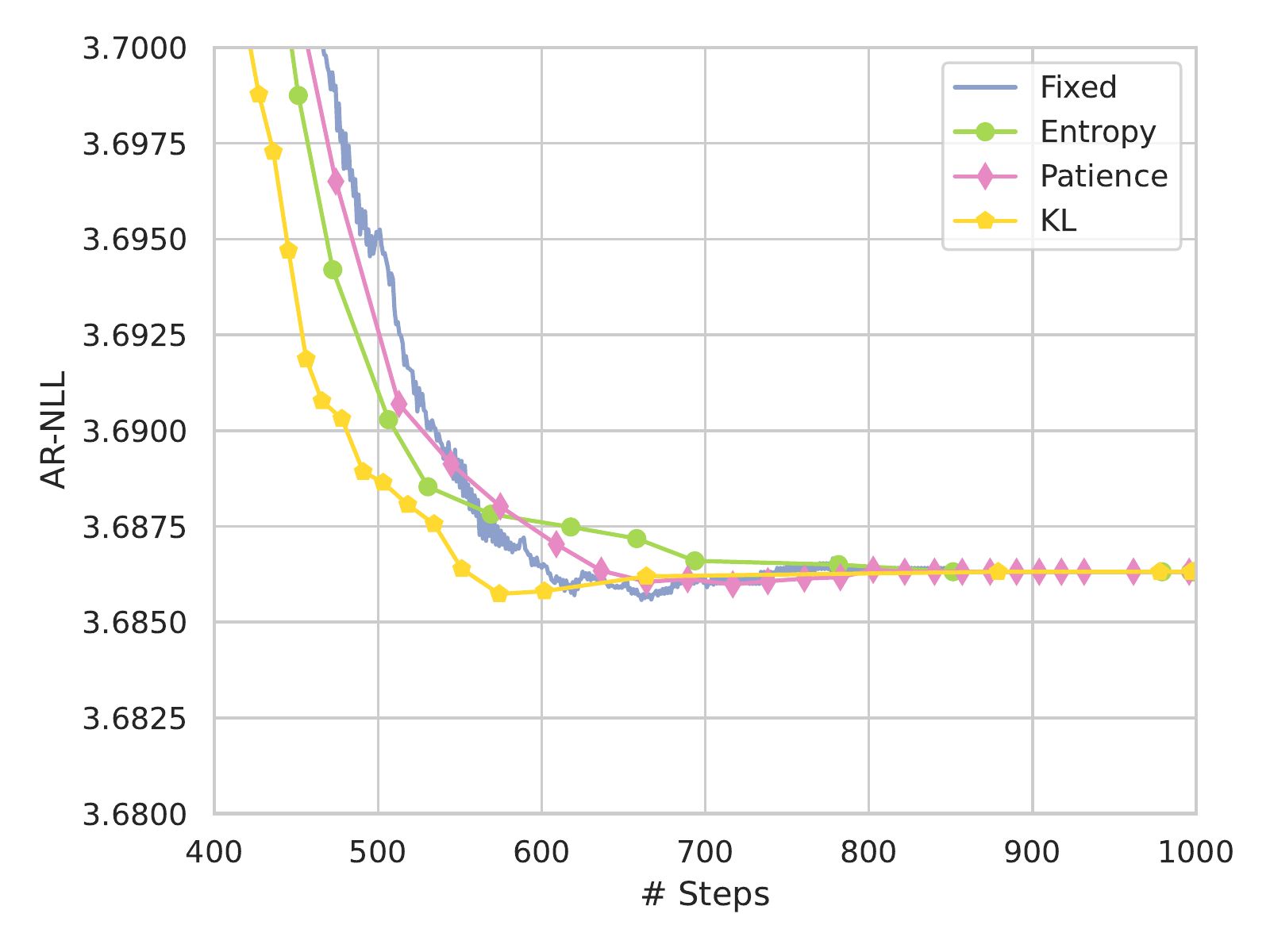}
    \caption{}
  \end{subfigure}
      \medskip
        \begin{subfigure}[t]{.32\linewidth}
    \centering\includegraphics[width=\linewidth]{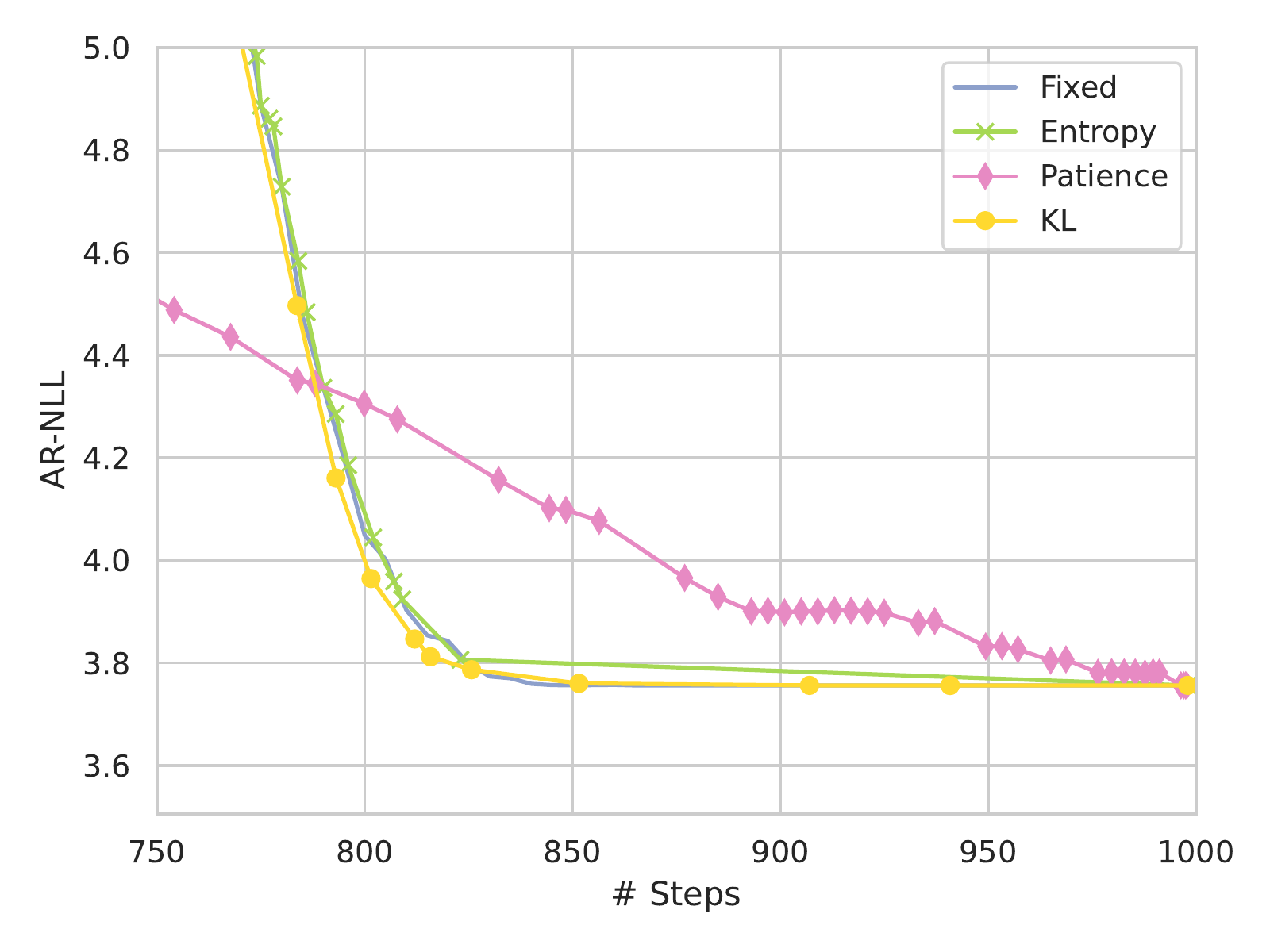}
    \caption{}
  \end{subfigure}
          \begin{subfigure}[t]{.32\linewidth}
    \centering\includegraphics[width=\linewidth]{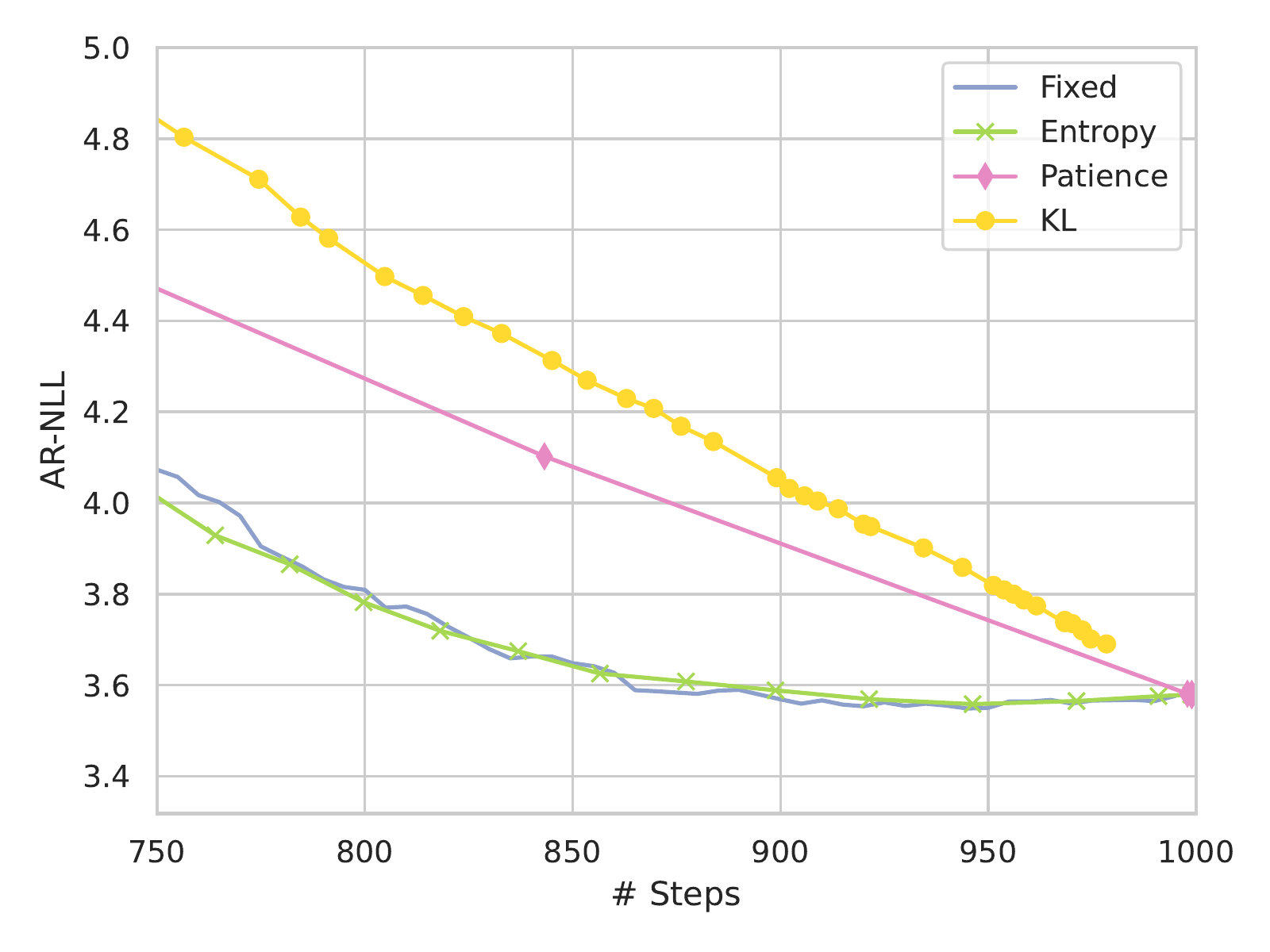}
    \caption{}
  \end{subfigure}


    \caption{(a) AR-NLL for the different exit criteria with DDLM, (b) SSD, and (c) Plaid with 1k samples of the C4 validation set. Our research shows that DDLM can effectively use adaptive early exiting strategies after step $600$, with the KL criterion allowing an exit $50$ steps earlier than other criteria (including fixed step exit) without loss in quality. The SSD model benefits modestly, with early exits saving about $10$ steps compared to different criteria and exiting after $850$-th step. Plaid lacks adaptive exiting effectiveness, with fixed criteria suggesting possible stops after step $900$ for computational efficiency. Overall, these approaches speed up text generation by up to $40$\% for DDLM, $10$-$15$\% for SSD, and $10$\% for Plaid, enhancing generation speed or sample quality. Despite differences in exit strategies, sample diversity remains unaffected, as indicated in Figure \ref{fig:uniq}. See Section \ref{section:dots} for more details.}
    \label{fig:aot}
\end{figure*}

We evaluated various early exiting criteria (described in Section \ref{section:method}) with DLMs. As seen in Figure \ref{fig:mean-criteria}, all the adaptive criteria applied to DDLM show that it may be possible to halt sampling during generation. For SSD, these criteria suggest stopping after the 800th step out of 1000. On the other hand, for Plaid, we observed that entropy decayed linearly during generation while other criteria remained constant. This suggests the possibility of Plaid performing poorly with adaptive early exiting methods.

\begin{figure*}

    \medskip
        \begin{subfigure}[t]{.32\linewidth}
    \centering\includegraphics[width=\linewidth]{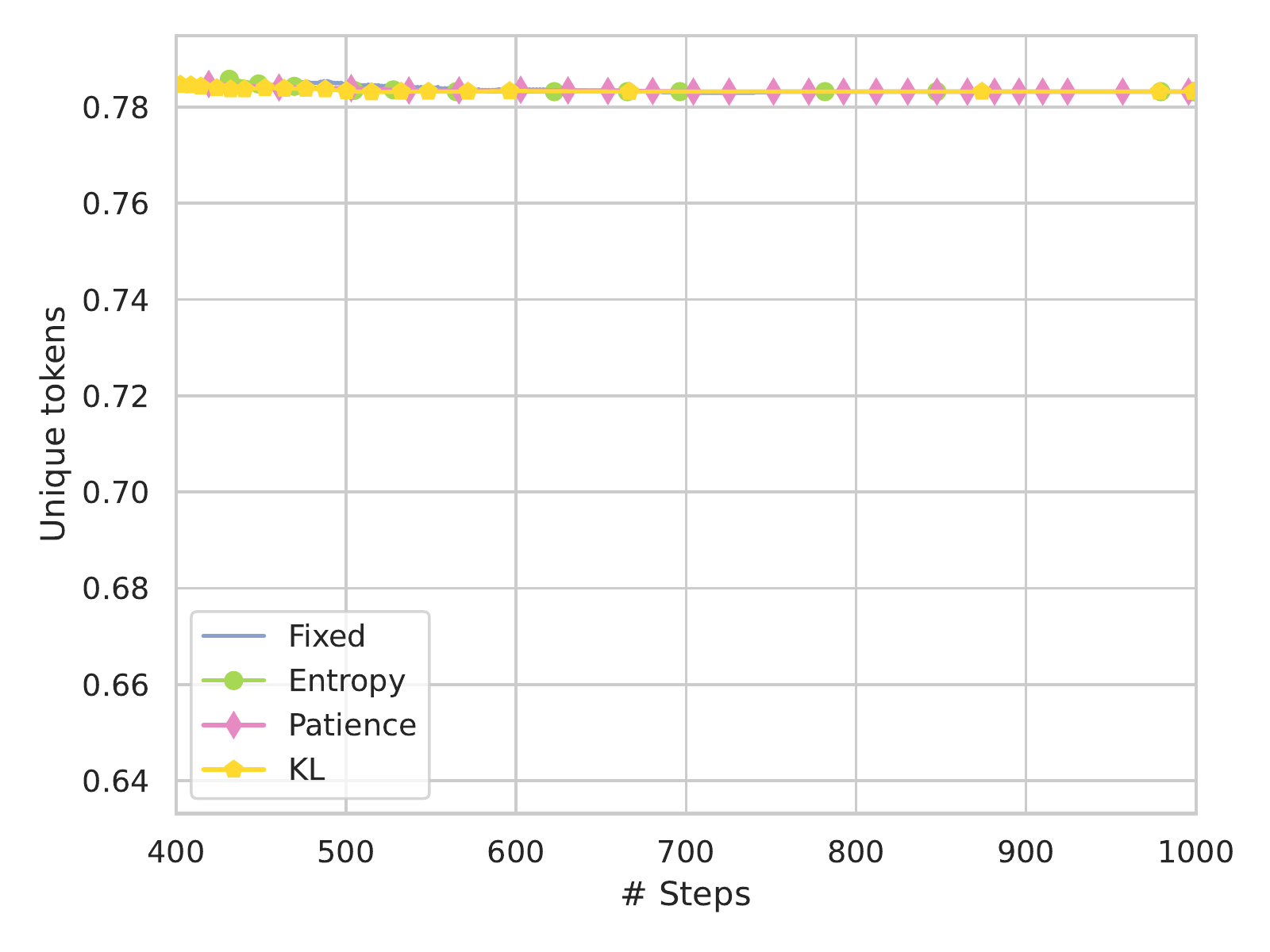}
    \caption{}
  \end{subfigure}
      \medskip
        \begin{subfigure}[t]{.32\linewidth}
    \centering\includegraphics[width=\linewidth]{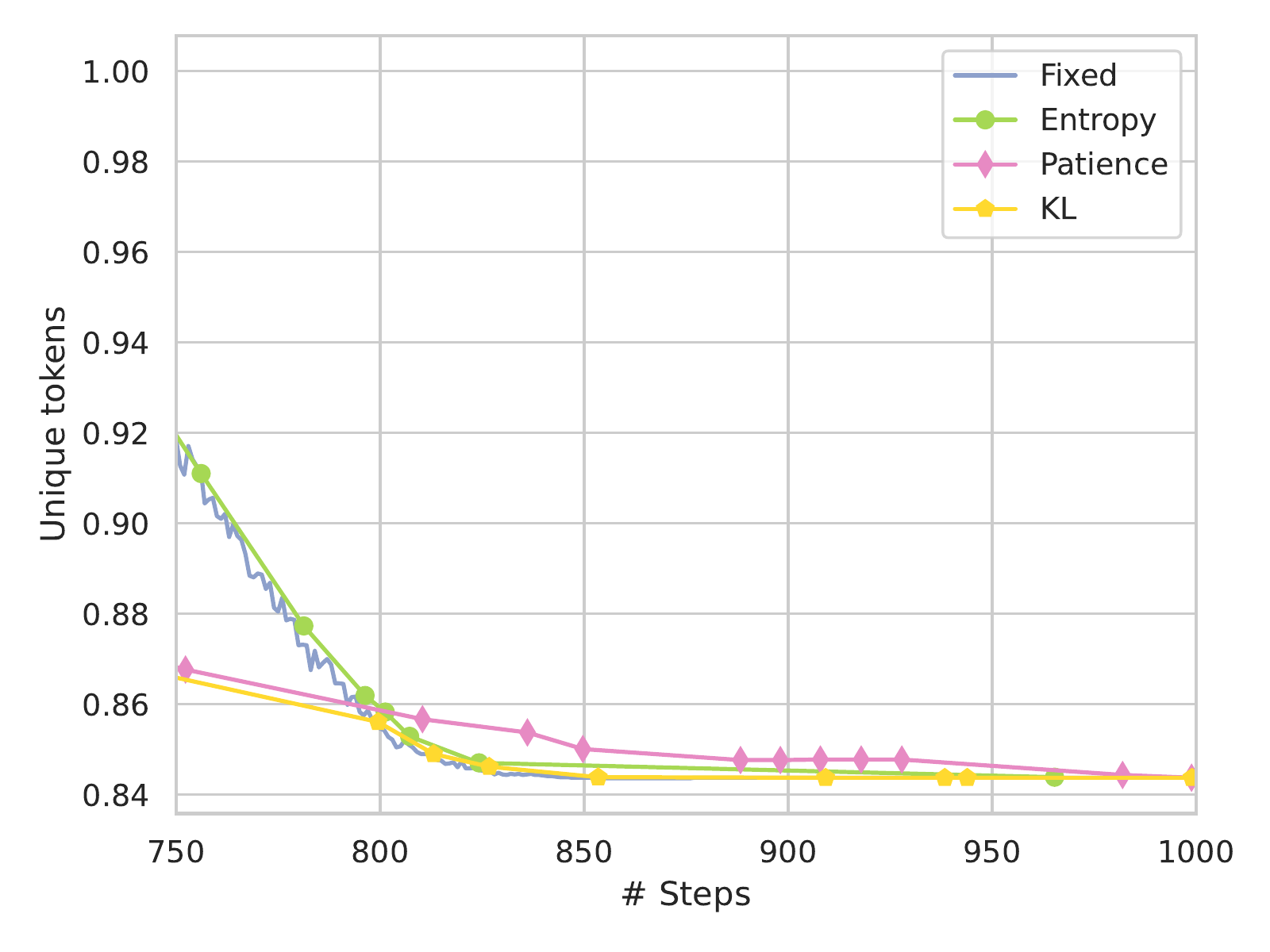}
    \caption{}
  \end{subfigure}
          \begin{subfigure}[t]{.32\linewidth}
    \centering\includegraphics[width=\linewidth]{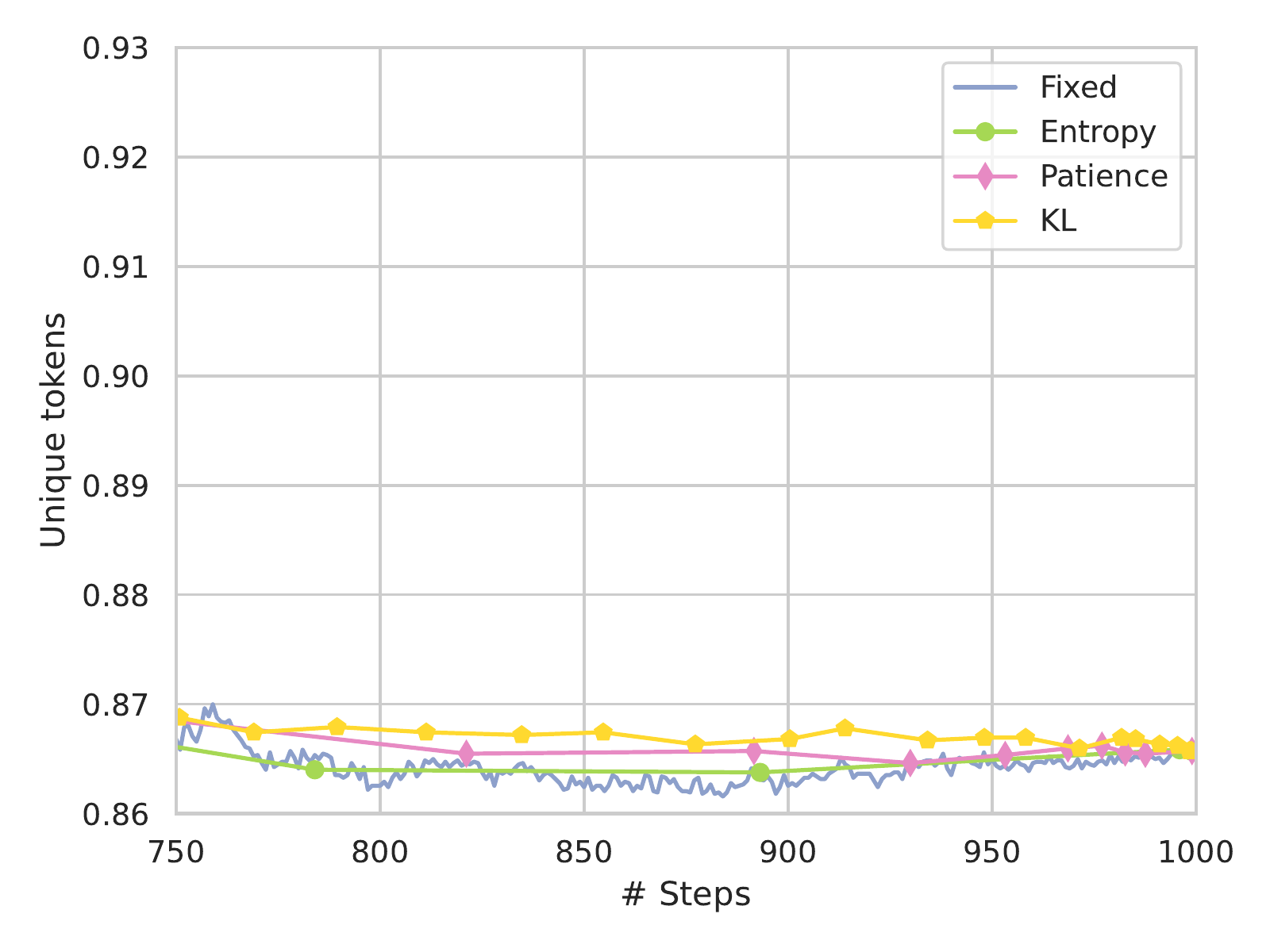}
    \caption{}
  \end{subfigure}


    \caption{Fraction of unique tokens for the different exit criteria with (a) DDLM, (b) SSD, and (c) Plaid with 1k samples of the C4 validation set. Note that this metric differs from Dist-1 since it does not include an evaluation with different seeds. We observed that no criteria reduce the number of unique tokens in generated samples. See Section \ref{section:dots} for more details.}
    \label{fig:uniq}
\end{figure*}

\subsection{Optimal Number of Steps} 
\label{section:dots}

In this experiment, we compared various adaptive early exiting criteria to a fixed strategy in optimizing generation steps across three models: DDLM, Plaid, and SSD. Our goal was to find the optimal threshold where we could reduce the mean number of generation steps without compromising sample quality.

We set up experiments using a Prefix-32 configuration and assessed quality by measuring the AR-NLL at each of the $1000$ generation steps. According to our results from Section \ref{section:method}, we anticipated DDLM's early exit around the $600$th step, SSD's adaptive exit after step $800$, and no adaptive exit for Plaid. However, it may be possible for Plaid to perform early exiting with a fixed exit step.

See Figure \ref{fig:aot} for results. We confirmed that DDLM and SSD could exit earlier using the KL criterion, achieving a step reduction and maintaining quality, while Plaid showed no benefit from adaptive strategies. Overall, our results show a speed increase of $40$\% for DDLM, $10$-$15$\% for SSD, and $10$\% for Plaid. This enables us either to generate text faster or improve text quality by allowing more steps in the same time frame. We also observed that early exiting methods do not hurt the diversity of samples (see Figure \ref{fig:uniq}).\footnote{One may find this result to contradict one observed with Section \ref{section:early_exit} and Table \ref{table:ranges} However, for experiments with noise scales, reduced variability is observed for small initial noise scales, leading to deterministic generation. At the same time, a noise scale equal to $1.0$ produces diverse samples, while early exiting methods do not hurt this variability.} See Appendix Figure \ref{fig:dots-256} for results with samples of length $256$.

\subsection{On Convergence of Early Exiting Methods}

We evaluate the sample dynamics during generation with GPT-4 \citep{gpt4} to understand the sample dynamics during generation. Recently, \citet{dpo} showed that this approach is comparable to human judgment and helps assess many samples for different time steps. We also calculate the Word Error Rate (WER) score between samples during generation and the sample from the final step. 

With such side-by-side assessment, our end goal is to understand the convergence of generations. GPT-4 allows us to compare samples with reference texts by considering their semantics, thus providing a broader evaluation. Meanwhile, WER shows the differences at the word level. See Appendix Section \ref{section:gpt-score} for more details on GPT-Score. 

\begin{figure}

    \medskip
    \medskip
        \begin{subfigure}[t]{.49\linewidth}
    \centering\includegraphics[width=\linewidth]{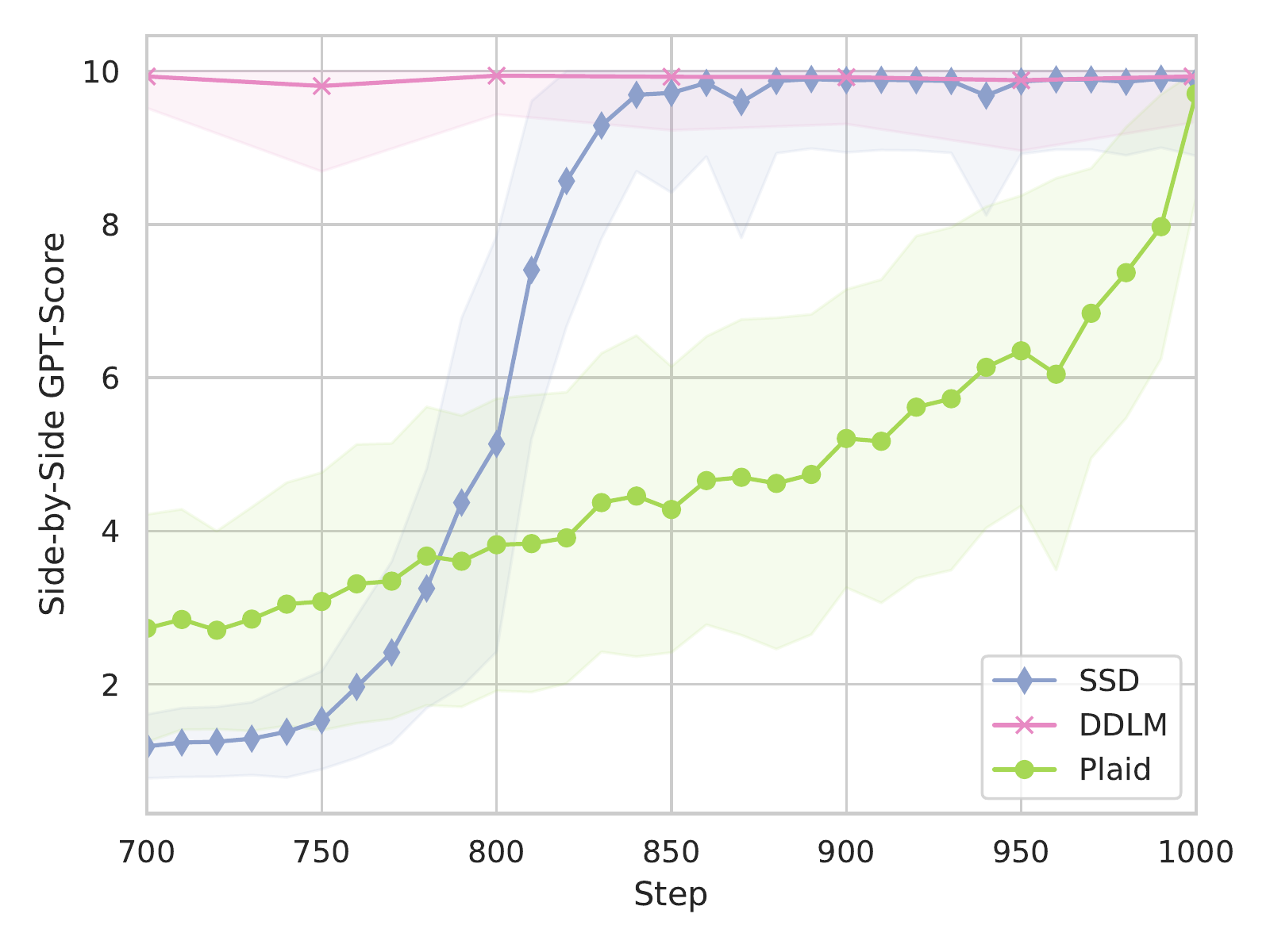}
    \caption{}
  \end{subfigure}
      \medskip
        \begin{subfigure}[t]{.49\linewidth}
    \centering\includegraphics[width=\linewidth]{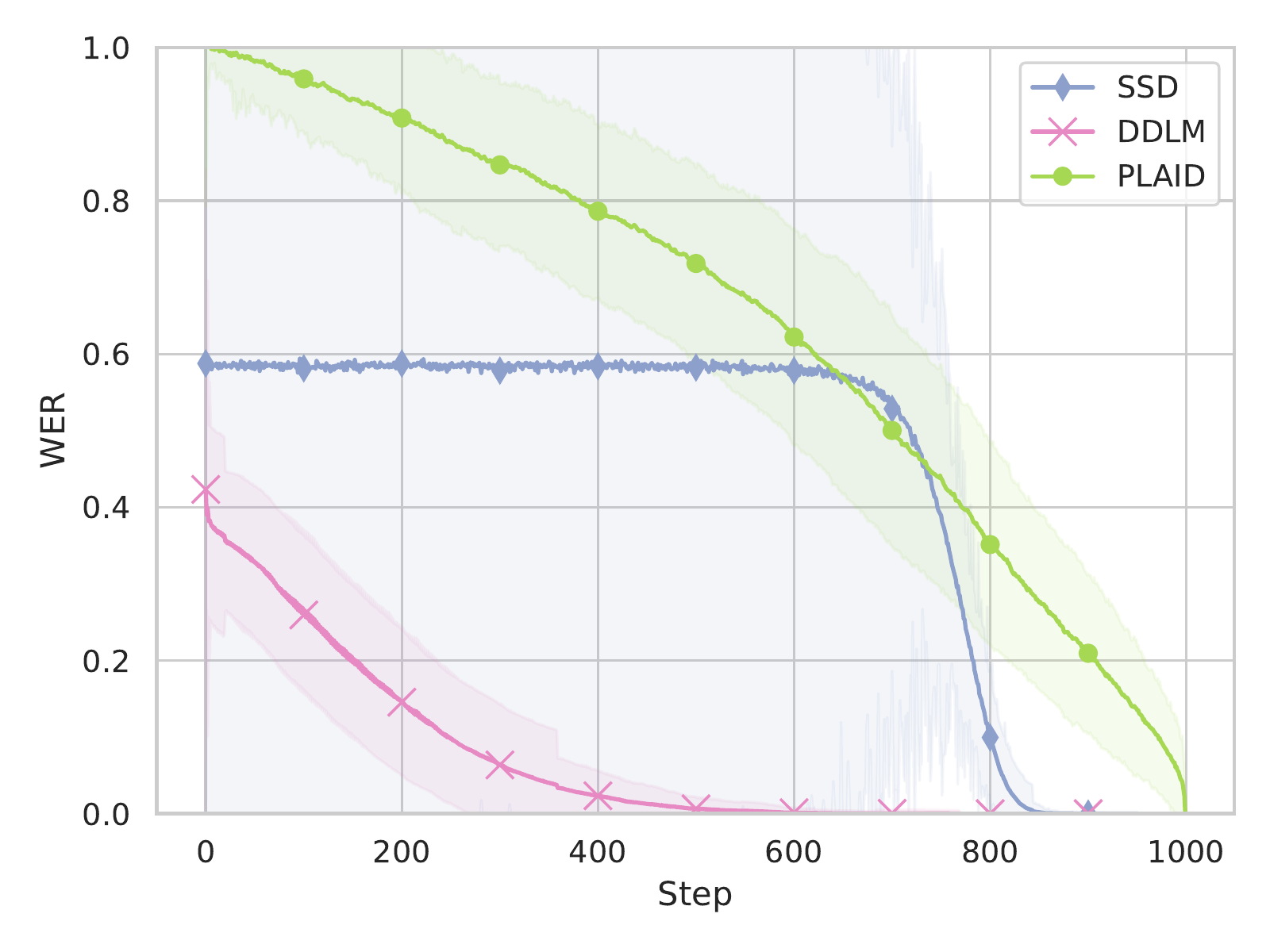}
    \caption{}
  \end{subfigure}

    \caption{(a) Side-by-side GPT-Score and (b) WER with final sample for DDLM, SSD, and Plaid models with a fixed early exiting mechanism. The plot is truncated to $400$ generation steps for GPT-Score. DDLM stabilizes at step $600$, SSD at $850$, and Plaid continues evolving until the end. However, after step $900$, Plaid shows minimal WER differences. For further information, refer to Section \ref{section:dots}.}
    \label{fig:sbs}
\end{figure}

Our results are presented in Figure \ref{fig:sbs}. DDLM converged with GPT-Score after the $600$th step, and there was no variance in samples afterward. For SSD, we observed the same behavior after the $850$th step. Meanwhile, for Plaid, we did not observe any convergence after the $900$th step with GPT-Score, and the GPT-Score of the side-by-side comparison with the final sample was large enough. The GPT-4 response indicated minor differences with the reference text, while WER reached low values, indicating that a fixed early exit could still be performed despite entropy not reaching its minimum. See Appendix Section \ref{section:samples} for sample examples.

\section{Discussion}

\textbf{Early Exiting Strategies.} One notable observation is that for both CDCD and SSD models, we can effectively implement adaptive techniques that allow the generation process to stop prematurely. In contrast, the Plaid model can halt generation without such adaptiveness. Most importantly, employing these early exiting tactics does not result in a decline in the generated content quality.

This finding has dual benefits. It can quicken text generation without quality loss or increase generation steps within a fixed timeframe to improve output quality. These enhancements promise broader adoption and ongoing advancement of DLMs.

\textbf{Identifying Issues in DLMs.} The ability to stop the text generation process early also signals opportunities to refine DLM design. We contemplate two possibilities: a) varying computational needs for different text generation tasks suggest early halting is apt for simple texts to prevent over-processing and beneficial for complex texts for additional computation; b) the computational effort may not vary with text complexity, suggesting that the capacity for early halting could point to design inefficiencies in DLMs (i.e., early exiting should not occur for properly trained and designed DLMs, thus indicating on issues with existing models). In the latter case, if the emergence of an early exit is an issue in the design of current DLMs, our research is a valuable methodology tool to evaluate and probe the performance of new pre-trained models. 

Considering dynamic generation processes is vital for deeply understanding model capabilities and their constraints. Such dynamic evaluations are often overlooked, with many studies preferring to assess a model's static performance using metrics like data likelihood \citep{plaid}. However, lessons from the Computer Vision field show that examining process dynamics can yield rich insights into specific cases \citep{Karras2023AnalyzingAI}.

\textbf{Directions for Future Research.} Our methodology offers insights into assessing the performance of emerging DLMs, noting that the option for early exiting could indicate underlying issues in the trained models. Therefore, future investigations could build upon our approach, incorporating new evaluation criteria or exploring DLMs that do not support early exiting. This could shed more detail on the strengths and potential weaknesses of these models.

\section{Limitations}

This paper only used our re-implementation of DLM trained with the CDCD framework, SSD, and Plaid models. We omitted other diffusion language models, such as GENIE or DiffuSeq  \citep{genie, diffuseq}, since there is no evidence that these frameworks can perform unconditional text generation if trained in such a manner. 

Our experiments involve our own DDLM model, which a reproduction of DLM trained with the CDCD framework. It is not a precise reproduction, as there is no source code available for CDCD. Nevertheless, we believe that conducting experiments on our model made it possible for us to present more comprehensive results in this paper.

Our analysis focuses on the AR-NLL metric to evaluate models during generation. However, our evaluation with GPT-4 indicates that no issues with the analysis should have occurred, and baseline models converged during generation.

\bibliography{custom}
\bibliographystyle{acl_natbib}

\newpage
\appendix
\onecolumn

\newcommand{\tokcur}{\mathrm{tokens}_\mathrm{cur}}
\newcommand{\tokprev}{\mathrm{tokens}_\mathrm{prev}}

\usebox{0}{
    \begin{minipage}{0.45\textwidth}
    \begin{algorithm}[H]    
    \caption{Entropy algorithm}\label{alg:e}
    \begin{algorithmic}[1]
        \Require{Diffusion model $f_\theta\left(\cdot, \cdot\right)$, entropy threshold $e_t$, maximum number of diffusion steps $N_\mathrm{max}$, timestamps array $t$.}
        \State $\mathrm{step} \gets 0$
        \State $x \gets X \sim \ \mathcal{N}(0, I)$
        \While{$\mathrm{step} < N_\mathrm{max}$}
        \State $p(\tokcur), \hat{x} \gets f_\theta(x, t\left[\mathrm{step}\right])$
        \State $e \gets \mathrm{entropy}(p(\tokcur))$
        \If{$e \leq e_t$}
            \State \textbf{return} $p(\tokcur)$
        \EndIf
        \State $x \gets$ Euler$(x, \hat{x}, t)$
        \State $\mathrm{step} \gets \mathrm{step} + 1$
        \EndWhile
        \State \textbf{return} $p(\tokcur)$
    \end{algorithmic}
    
    \end{algorithm}
    \end{minipage}
}
\hfill
\usebox{0}{
    \begin{minipage}{0.45\textwidth}
    \begin{algorithm}[H]    
    \caption{Patience algorithm}\label{alg:p}
    \begin{algorithmic}[1]
        \Require Diffusion model $f_\theta(\cdot, \cdot)$, patience threshold $p$, maximum number of diffusion steps $N_\mathrm{max}$, timestamps array $t$
        \State $\mathrm{step} \gets 0$
        \State $p_\mathrm{cur} \gets 0$
        \State $x \gets X \sim \ \mathcal{N}(0, I)$
        \While{$\mathrm{step} < N_\mathrm{max}$}
        \State $p(\tokcur), \hat{x} \gets f_\theta(x, t[\mathrm{step}])$ 
        \State $\tokcur \gets$ argmax$(p(\tokcur))$
        \If{$\mathrm{step}$ > 0}
            \If{$\tokcur = \tokprev$}
                \State $p_\mathrm{cur} \gets p_\mathrm{cur} + 1$
            \Else
                \State $p_\mathrm{cur} \gets 0$
            \EndIf
            \If{$p_\mathrm{cur} \geq p$}
               \State \textbf{return} $p(\tokcur)$
            \EndIf
        \EndIf
        \State $x \gets$ Euler$(x, \hat{x}, t)$
        \State $\tokprev \gets \tokcur$
        \State $\mathrm{step} \gets \mathrm{step} + 1$
        \EndWhile
        \State \textbf{return} $p(\tokcur)$
    \end{algorithmic}
    \end{algorithm}
    \end{minipage}
}
\begin{algorithm}[H]    
    \caption{KL algorithm}\label{alg:kl}
    \begin{algorithmic}[1]
        \Require{Diffusion model $f_\theta(\cdot, \cdot)$, divergence-threshold $d$, maximum number of diffusion steps $N_\mathrm{max}$,  parameter $\mathrm{min\_steps} \approx 0.25 N_\mathrm{max}$, timestamps array $t$, }
        \State $\mathrm{step} \gets 0$
        \State $x \gets X \sim \ \mathcal{N}(0, I)$
        \While{$\mathrm{step} < N_\mathrm{max}$}
        \State $p(\tokcur), \hat{x} \gets f_\theta(x, t\left[\mathrm{step}\right])$
        \If{$\mathcal{D}((p(\tokcur)||p(\tokprev)) > d_t$ \textbf{and} $s \geq \mathrm{min\_steps}$}
            \State \textbf{return} $p(\tokcur)$
        \EndIf
        \State $x \gets$ Euler$(x, \hat{x}, t)$
        \State $\mathrm{step} \gets \mathrm{step} + 1$
        \State $p(\tokprev) \gets p(\tokcur)$
        \EndWhile
        \State \textbf{return} $p(\tokcur)$
    \end{algorithmic}
    \end{algorithm}

\begin{figure*}

    \medskip
        \begin{subfigure}[t]{.45\linewidth}
    \centering\includegraphics[width=\linewidth]{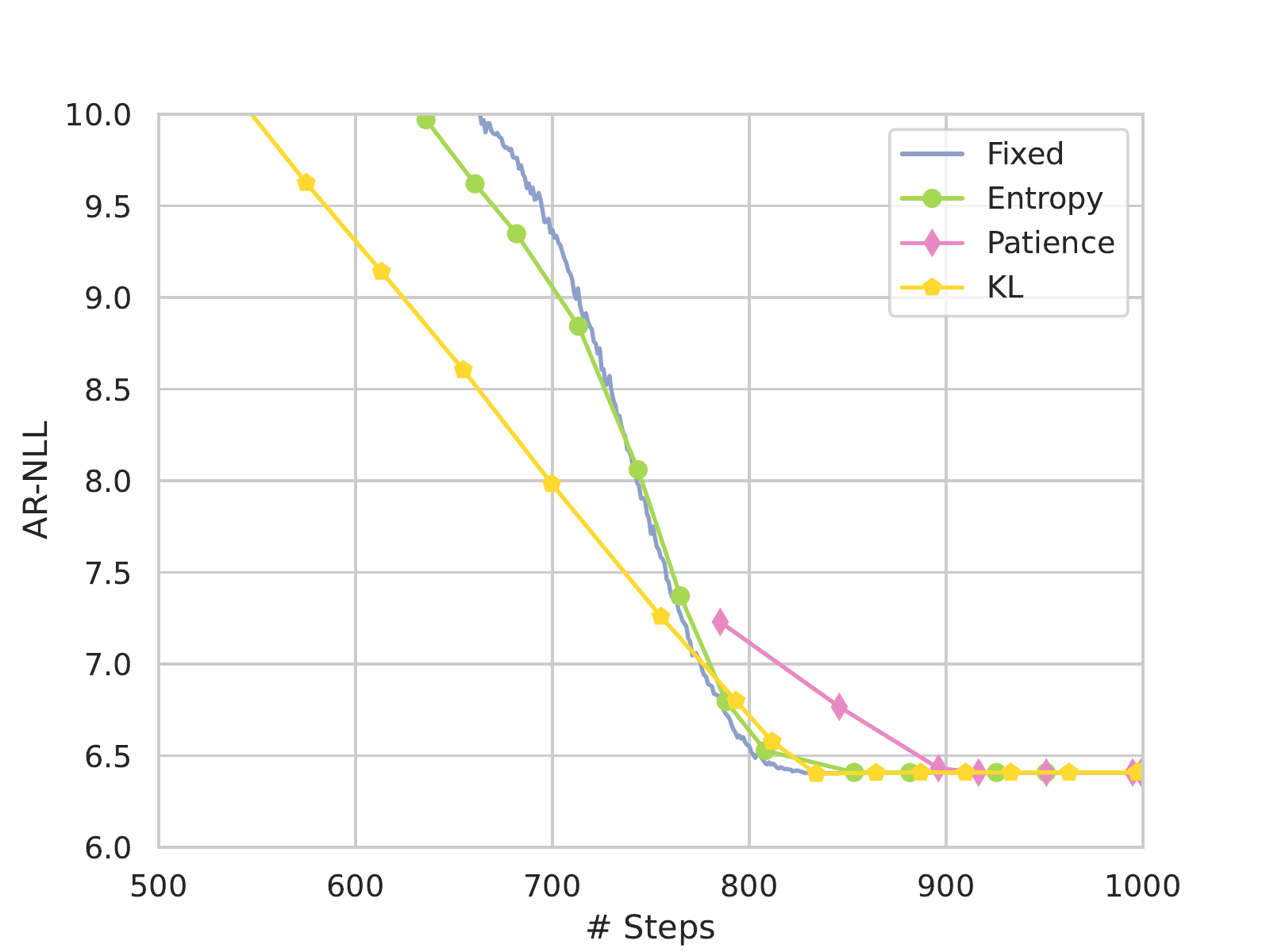}
    \caption{}
  \end{subfigure}
      \medskip
        \begin{subfigure}[t]{.45\linewidth}
    \centering\includegraphics[width=\linewidth]{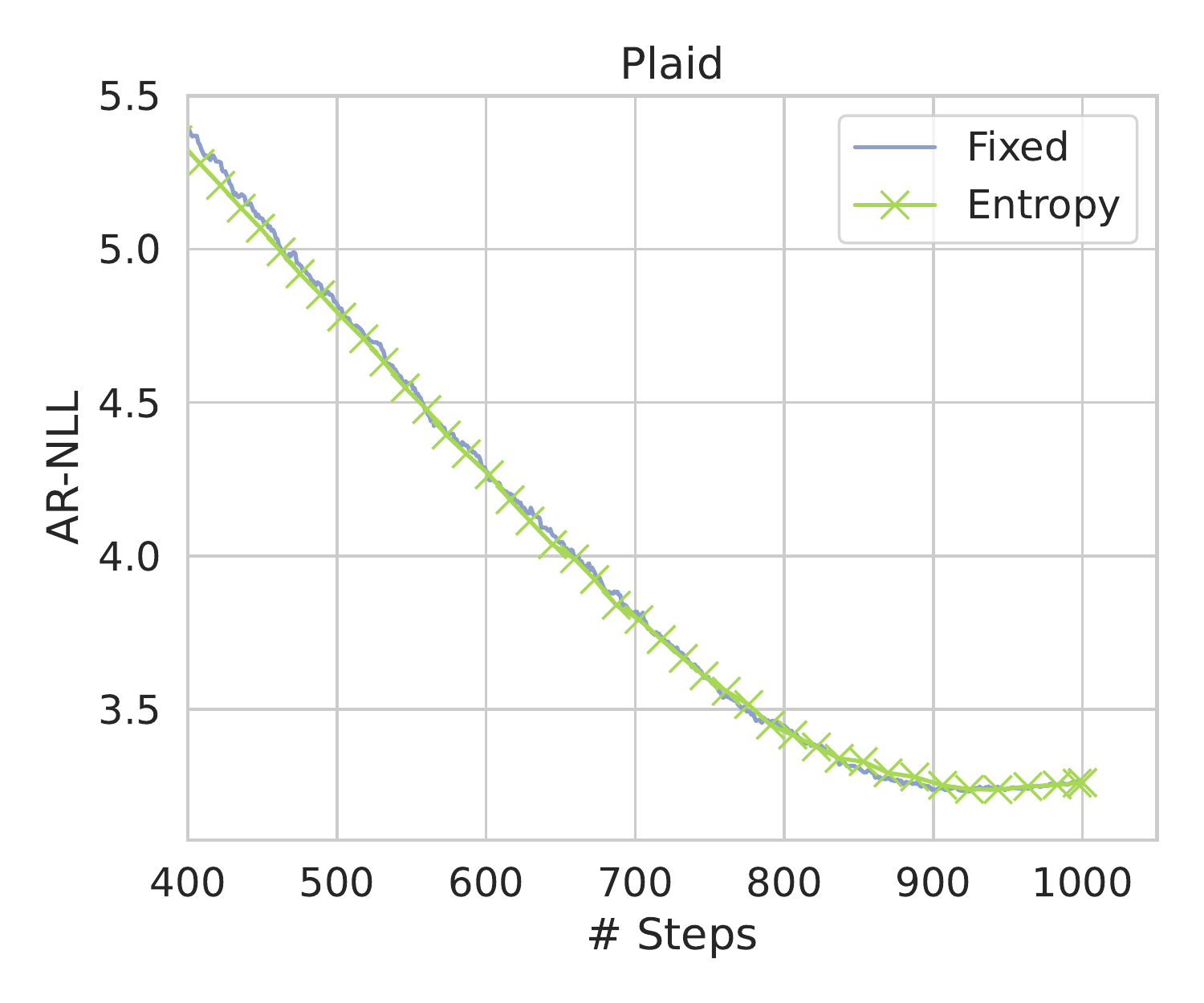}
    \caption{}
  \end{subfigure}


    \caption{(a) AR-NLL of samples with length $256$ for the different exit criteria with SSD, and (b) Plaid with 200 samples of the C4 validation set. Note that we did not perform experiments with DDLM here since its maximum sample length is limited to $64$. Early exiting behavior remains with longer sequences. See Section \ref{section:dots} for more details.}
    \label{fig:dots-256}
\end{figure*}

\begin{table*}[ht!]
\centering

\small
\begin{tabular}{ccccccc}
\toprule
L    & H              & D      & Seq. len.  & Masking     & Optim. & Time Warping \\

  8   &        8        &    1024    & 64         & [MLM, Prefix, Span]         & Adam & [no, yes]   \\
\midrule
LR   & Scheduler      & Warmup & Batch size & $t_{max}$ & Steps &   \\
3e-5 & Cos. w/ Warmup & 10k     & 1024       & [10, 50, 300]        & 1e6 &    \\
\toprule
\end{tabular}
\caption{Pre-training hyperparameters used for experiments with noise scheduling (See Section \ref{section:ddlm-training}). L stands for number of layers, H for number of heads in the Transformer layer, and D for hidden size.}
\label{tab:hyp_pre_t}
\end{table*}

\begin{table}[ht!]
    \begin{center}
    \begin{small}
\begin{tabular}{r|c|c|cccccc}
\toprule
Model                  & Steps & Sampler & AR-NLL        & Dist-1        & Dist-2        & Dist-3        & MAUVE         & Zipf's Coef.  \\
\midrule
Data                   & N/A   & N/A     & 3.31          & N/A           & N/A           & N/A           & N/A           & 0.90          \\
\midrule
\multicolumn{9}{c}{Prefix-32}                                                                                                            \\
\midrule
\multirow{3}{*}{DDLM, 147M}  & 50    & \multirow{3}{*}{Euler}   & 3.72          & 0.53          & 0.85          & \textbf{0.90}  & 0.80          & 0.96          \\
                       & 200   &    & 3.65          & 0.54          & 0.84          & \textbf{0.90}  & 0.82          & 0.96          \\
                       & 1000  &    & \textbf{3.63} & 0.54          & 0.84          & \textbf{0.90}  & 0.81          & 0.96          \\
                       \midrule
\multirow{3}{*}{Plaid, 1.3B} & 200   & \multirow{3}{*}{DDPM}    & 3.69          & \textbf{0.66} & 0.88          & \textbf{0.90}  & 0.93          & 0.86          \\
                       & 500   &     & 3.64          & 0.65          & 0.87          & 0.89          & 0.89          & 0.87          \\
                       & 1000  &     & 3.65          & 0.65          & 0.87          & \textbf{0.90}  & \textbf{0.94} & 0.87          \\
                       \midrule
\multirow{2}{*}{SSD, 400M}   & 200   &   \multirow{2}{*}{Simplex}      & 4.00           & \textbf{0.66}          & \textbf{0.91} & 0.83          & 0.82          & 0.88          \\
                       & 1000  &         & 3.75          & 0.63          & \textbf{0.91} & 0.83          & 0.85          & \textbf{0.90} \\
                       \midrule
                       GPT-2, 124M                  & N/A   & N/A     & 3.21          & 0.58          & 0.86          & 0.89          & 0.86          & 0.96          \\
GPT-Neo, 125M                & N/A   & N/A     & 3.20          & 0.60          & 0.85          & 0.88          & 0.83          & 0.96          \\
\midrule
\multicolumn{9}{c}{Unconditional}                                                                                                        \\
\midrule
\multirow{3}{*}{DDLM, 147M}  & 50    & \multirow{3}{*}{Euler}   & 3.98          & 0.50           & 0.85          & 0.93          & N/A           & 1.19          \\
                       & 200   &    & 3.77          & 0.50           & 0.84          & 0.92          & N/A           & 1.17          \\
                       & 1000  &    & \textbf{3.67} & 0.49          & 0.83          & 0.91          & N/A           & 1.16          \\
                       \midrule
\multirow{3}{*}{Plaid, 1.3B} & 200   & \multirow{3}{*}{DDPM}    & 3.83          & \textbf{0.66} & \textbf{0.92} & \textbf{0.94} & N/A           & \textbf{0.93} \\
                       & 500   &     & 3.73          & 0.65          & 0.91          & \textbf{0.94} & N/A           & 0.94          \\
                       & 1000  &     & 3.69          & 0.65          & 0.91          & \textbf{0.94} & N/A           & 0.94          \\
                       \midrule
\multirow{2}{*}{SSD, 400M}   & 200   &   \multirow{2}{*}{Simplex}      & 6.45          & 0.57          & 0.91          & 0.83          & N/A           & 0.99          \\
                       & 1000  &         & 6.55          & 0.57          & 0.91          & 0.83          & N/A           & 1.12         \\
                       \midrule
                       GPT-2, 124M                  & N/A   & N/A     & 2.62          & 0.67          & 0.90          & 0.90          & N/A           & 1.10          \\
GPT-Neo, 125M                & N/A   & N/A     & 2.27          & 0.66          & 0.88          & 0.89          & N/A           & 1.05   \\
                       \bottomrule
\end{tabular}
        \caption{Evaluation of DDLM, SSD, Plaid, GPT-2, and GPT-Neo with 5k samples of the C4 validation set with the Unconditional and Prefix-32 tasks. The best result across DLMs is bolded. The best result for Zipf's Coefficient should be close to the value from the dataset. See Section \ref{section:ddlm-training} for more details.}
            \label{tab:final}
    \end{small}
    \end{center}
\end{table}

\section{Reproducing CDCD}
\label{section:ddlm-training}

Comparing the CDCD model with other diffusion models is an intriguing challenge due to its unique objectives that set it apart from conventional DLMs. However, the lack of a publicly available training code for the CDCD limits such research. Therefore, we have reproduced this model in order to understand the differences between CDCD and other frameworks. We will briefly describe the essential parts of the CDCD framework and then go into detail about our reproduction of the CDCD.

\subsection{Understanding CDCD Framework}
\label{section:understanding}

Once loss and score functions are defined, CDCD implies several details must be considered before training a model.

The first of them is \textbf{Embeddings normalization}. As the model with the $\mathcal{L}_{\text{CDCD}}$ loss function is forced to distinguish correct embeddings from noisy ones, a naive application of such an objective will lead to uncontrollable growth of the embeddings norm to make them easier to distinguish. CDCD applies $L_2$ normalization during training to prevent an uncontrolled growth of embedding norms.

Second, the score interpolation objective implies sampling the time $t$ from some distribution during the training. While it is possible to sample $t$ uniformly in $[0; 1]$, \citet{continuous_diffusion} used \textbf{Time Warping} method. \citet{continuous_diffusion} trained CDF of time $\mathrm{F}_\phi(t)$ following \citet{variational_diffusion}. More concretely, for the CDCD framework, $\mathrm{F}_\phi(t)$ is trained with a loss $\mathcal{L}_{\text{TW}} = \| \widetilde{\mathrm{F}}_\phi(t) - \mathcal{L}_{\text{CDCD}}(\dots, t)\|$, where $\widetilde{\mathrm{F}}_\phi(t)$ is the unnormalized CDF parametrized with $\phi$. We can obtain samples from it by normalizing and inverting  $\widetilde{\mathrm{F}}_\phi(t)$. $p(\vx|\mX, t)$ is then conditioned on $t$ via conditional layer normalization \citep{film}. 

Finally, since our model is trained akin to Masked Language Models to fill noisy tokens with real ones, it is essential to define the mechanism to select specific tokens to inject noise, i.e., \textbf{Noise masking}. The first approach, prefix masking, involves injecting noise into the embedding sequence continuation while keeping its beginning intact. Alternatively, noise can be injected at random sequence positions, similar to Masked Language Models training (MLM masking) \citep{bert, deberta, roberta, albert}. The third approach combines the previous two, injecting noise into random positions in a sequence continuation (mixed masking). The cross-entropy loss $\mathcal{L}_{\text{CDCD}}$ is calculated only with noised embeddings.

CDCD is implemented as Transformer \citep{transformer}. Once all objective embeddings necessary for score interpolation are concatenated, they are passed through Transformer layers to obtain $p(\vx|\mX, t)$.

\subsection{Training DDLM}
Following the information provided on the CDCD framework, we trained our version of it, namely the Democratized Diffusion Language Model (DDLM).\footnote{"Democratized" in the model name stands for the open availability of this model for other researchers.} We trained this model using the C4 dataset \citep{t5}  with 147M parameter models and a sequence length of 64 tokens.

The tokenized training data consisted of a vocabulary $|V| = 32\text{k}$, and the tokens used 256-sized embeddings following \citet{continuous_diffusion}. We trained DDLM using 8 NVidia A100 SXM4 80GB GPUs, completing one million training steps over approximately 1.5 days. The details on the hyperparameters used can be found in Table \ref{tab:hyp_pre_t}.

For validation, we extracted 5k examples from the C4 validation set and generated 5 separate continuations using different seeds. Our evaluation of DDLM was carried out in two setups: Unconditional and Prefix-32, where text was generated using a prefixed prompt of 32 tokens in length.

While \citet{continuous_diffusion} states that small values of $t_{\text{max}}$ can lead to trivial solutions for the score interpolation objective, we hypothesize that applying several normalizations during training, such as normalizing embeddings and noised embeddings, can prevent trivial solutions from emerging.

Additionally, our interest extended to delving deeper into noise-masking strategies. While \citet{continuous_diffusion} favored mixed masking, we suggested an extension of prefix masking, a component of mixed masking, to span masking \citep{sed}. In span masking, a sequence of tokens is divided into $k$ segments ($k$ being a randomly chosen integer between $1$ and a fixed constant $k_{max}=9$) by randomly selecting $k-1$ indices. These indices define $k$ spans, each subjected to noise with a probability of 50\%. It is important to note that our experimentation with the span masking strategy was not aimed at achieving superior performance compared to other methods, but rather at uncovering their distinctions.

We trained models with different $t_{\text{max}}$ values, including $t_{\text{max}} \in [10, 50, 300]$. Both models with and without time warping were trained for each $t_{\text{max}}$ value. Furthermore, all these experiments were conducted using three masking strategies: MLM, prefix, and span.

For the detailed results of Unconditional, Prefix-32, and Enclosed-32 generation, refer to Table \ref{tab:ablation} and Appendix Tables \ref{tab:ablation-unconditional}, \ref{tab:ablation-prefix}, and \ref{tab:ablation-enclosed}. We observed that training models with high $t_{\text{max}}$ values led to poor results with repetitive samples. Comprehensive samples were only achieved when $t_{\text{max}}$ was reduced to $10$. Notably, while larger $t_{\text{max}}$ values resulted in poor samples, the loss values for such setups did not indicate inadequate training. This suggests that the loss values of Diffusion LMs trained with score interpolation should not be compared directly with those of other methods.

When comparing different training setups with $t_{\text{max}} = 10$, a model with the MLM masking strategy and time warping achieved the best AR-NLL score. The second-best model was trained with a Span masking strategy and no time warping. It is important to highlight that the slightly lower Dist-1 metric values of the first model might be linked to its lower AR-NLL score. Additionally, it is worth noting that prefix masking yielded inferior results compared to other masking strategies on the Enclosed-32 task. We can assume that this outcome can be attributed to the fact that, during pre-training, only left-conditioning was employed with this type of masking, restricting the model's ability to generate sequences conditioned from both sides.

In comparing these results with those reported by \citet{continuous_diffusion}, we observed a discrepancy in the best-performing noise scales due to the poor reproducibility of the original CDCD, which led to differences in CDCD and DDLM training pipelines. While the original CDCD evaluation used an unnamed language model (possibly proprietary), preventing direct comparison of the results (e.g., with the AR-NLL metric), the AR-NLL metrics reported by \citet{continuous_diffusion} are comparable to our results, even considering potential variations from using GPT-Neo-1.3B.

For the experiments, \textbf{we refer to DDLM as the model with MLM masking strategy, $t_{\text{max}} = 10$, and time warping.}

The evaluation results for our DDLM model are summarized in Table \ref{tab:final}. We observed that DDLM performs competitively when compared to Plaid in terms of AR-NLL values, although Plaid did excel at generating a larger number of distinct tokens across samples. The SSD model displayed comparable performance to DDLM and Plaid in the conditional generation setup, but demonstrated significantly higher AR-NLL values in the unconditional setup, indicating a weaker ability to model sequences in complex multimodal conditions \citep{nag}. Overall, all DLMs underperformed when compared to autoregressive LMs in terms of AR-NLL values.\footnote{This observation contradicts the findings of \citet{plaid}. However, it is worth noting that \citet{plaid} compared Plaid to GPT-2 based only on NLL values, without evaluating the generated sequences.}

\section{GPT-Score Details}
\label{section:gpt-score}
The instruction contained a request to evaluate a text's spelling, consistency, and coherence with a number from 1 to 10 compared to the sampling from the last $1000$-th generation step, which served as a reference. Also, we included requesting for ignoring abrupt endings of texts since all models were evaluated with sample length equal to $64$.

\begin{AIbox}{}
\textbf{\color{purple}System prompt:}\\ \tt{\color{purple}\footnotesize 
Act as a human annotator. Strictly follow the provided instructions. }
\\
\textbf{\color{purple}Instruction:}\\ \tt{\color{purple}\footnotesize 
Evaluate the quality of the provided text compared to the reference text}.

Text: 'text'

Reference: 'reference'

FIRST, provide a one-sentence explanation of issues in the provided text compared to the reference text.

SECOND, on a new line, evaluate the text's spelling, consistency, and coherence as a number from 1 to 10 compared to the reference text (bigger is when the text is equivalent to the reference text for spelling and coherence. Use 10 only if provided text is better than reference text). Ignore abrupt endings of texts. Note that the first half of the text is provided as a prompt. Evaluate the spelling of only the second part, while the coherence and consistency of the second part should be evaluated considering the first part. 

Your response should use the format:

Explanation: <one-sentence explanation>

Result: <a number from 1 to 10>

\end{AIbox}

\section{Sample Examples}
\label{section:samples}
We report samples from each model from different generation steps. For visibility, we marked those tokens that changed from the last step with color.

\subsection{DDLM}

\begin{AIbox}{}
\tt{\footnotesize  \textbf{\color{purple} Step: 0:}\\ASHION was \textcolor{purple}{.} \textcolor{purple}{ for} \textcolor{purple}{ a} \textcolor{purple}{ business} \textcolor{purple}{ of} \textcolor{purple}{.} is \textcolor{purple}{ date} of \textcolor{purple}{ that} \textcolor{purple}{ this} registered \textcolor{purple}{ of} the \textcolor{purple}{
} \textcolor{purple}{.} \textcolor{purple}{
} registered \textcolor{purple}{ information} \textcolor{purple}{ for} \textcolor{purple}{ F}. \textcolor{purple}{ company} \textcolor{purple}{.,} \textcolor{purple}{EL}\\\textbf{\color{purple} Step: 250:}\\ASHION was born 24 January 18 \textcolor{purple}{.} and is \textcolor{purple}{ the} of \textcolor{purple}{ female}, registered from the New \textcolor{purple}{,} voter registered as of \textcolor{purple}{ 1972}.
CH \textcolor{purple}{EN}\\\textbf{\color{purple} Step: 500:}\\ASHION was born 24 January 1896 and is part of Florida, registered from the New York voter registered as of 2019.
CHD\\\textbf{\color{purple} Step: 999:}\\ASHION was born 24 January 1896 and is part of Florida, registered from the New York voter registered as of 2019.
CHD\\
}
\end{AIbox}

\begin{AIbox}{}
\tt{\footnotesize  \textbf{\color{purple} Step: 0:}\\ it \textcolor{purple}{ people} \textcolor{purple}{ was}, the \textcolor{purple}{ the} \textcolor{purple}{,'} \textcolor{purple}{s} is \textcolor{purple}{,} the \textcolor{purple}{'s} of the \textcolor{purple}{,} \textcolor{purple}{."} \textcolor{purple}{ in} \textcolor{purple}{ he} \textcolor{purple}{ success}.
 \textcolor{purple}{,"} \textcolor{purple}{H} \textcolor{purple}{ that} \textcolor{purple}{ are} a \textcolor{purple}{ B} \textcolor{purple}{ to}, \textcolor{purple}{ is} \textcolor{purple}{ B}\\\textbf{\color{purple} Step: 250:}\\ it turns out, the old \textcolor{purple}{ football} \textcolor{purple}{ ball} is about the price of the ball," \textcolor{purple}{,"} \textcolor{purple}{ W} said.
"If you have a good \textcolor{purple}{ one}, you're\\\textbf{\color{purple} Step: 500:}\\ it turns out, the old sports game is about the price of the ball," Berley said.
"If you have a good shot, you're\\\\
}
\end{AIbox}

\subsection{SSD}
\begin{AIbox}{}
\tt{\footnotesize  
\textbf{\color{purple} Step: 0:}\\<s> As \textcolor{purple}{ utility} \textcolor{purple}{rent} \textcolor{purple}{ wood} \textcolor{purple}{ights} \textcolor{purple}{ releases} \textcolor{purple}{ oblivious} \textcolor{purple}{ incent} \textcolor{purple}{ signature} \textcolor{purple}{ infusion} \textcolor{purple}{ Maine} \textcolor{purple}{B} \textcolor{purple}{ult} \textcolor{purple}{ested} \textcolor{purple}{Throw} \textcolor{purple}{ cloth} \textcolor{purple}{0000000000000000} \textcolor{purple}{ Serve} \textcolor{purple}{ floated} \textcolor{purple}{ q} \textcolor{purple}{ lives} \textcolor{purple}{ depleted} \textcolor{purple}{acked} \textcolor{purple}{ conduct} \textcolor{purple}{ Tina} \textcolor{purple}{ catchy}\\\textbf{\color{purple} Step: 250:}\\<s> As \textcolor{purple}{Individual} \textcolor{purple}{ Ashes} \textcolor{purple}{ Waterloo} \textcolor{purple}{ Marshal} \textcolor{purple}{ set} \textcolor{purple}{ Allen} \textcolor{purple}{ Mission} \textcolor{purple}{ incremental} \textcolor{purple}{ Bac} \textcolor{purple}{ 110} \textcolor{purple}{ustainable} \textcolor{purple}{ Hearth} \textcolor{purple}{ENCE} \textcolor{purple}{ Micro} \textcolor{purple}{ Kislyak} \textcolor{purple}{amber} \textcolor{purple}{ unconsciously} \textcolor{purple}{ Naval} \textcolor{purple}{ topp} \textcolor{purple}{ Ratings} \textcolor{purple}{ gob} \textcolor{purple}{ tariff} \textcolor{purple}{ss} \textcolor{purple}{usp} \textcolor{purple}{ reinforcing} \textcolor{purple}{ mammalian}\\\textbf{\color{purple} Step: 500:}\\<s> As \textcolor{purple}{ foreseeable} \textcolor{purple}{','} \textcolor{purple}{ vote} \textcolor{purple}{ Song} \textcolor{purple}{ withdrawal} \textcolor{purple}{ (} \textcolor{purple}{ Thro} \textcolor{purple}{ sang} \textcolor{purple}{severe} \textcolor{purple}{Were} \textcolor{purple}{Taylor} \textcolor{purple}{ Grill} \textcolor{purple}{ Johns} \textcolor{purple}{atus} \textcolor{purple}{ anarchists} \textcolor{purple}{ ][} \textcolor{purple}{ pressures} \textcolor{purple}{ournament} \textcolor{purple}{ Taiwan} \textcolor{purple}{ believable} \textcolor{purple}{zens} \textcolor{purple}{ squad} \textcolor{purple}{Eth} \textcolor{purple}{ its} \textcolor{purple}{290} \textcolor{purple}{ dont}\\\textbf{\color{purple} Step: 750:}\\<s> As one of the \textcolor{purple}{ semin} \textcolor{purple}{Merc} Boc \textcolor{purple}{ 450} Ball \textcolor{purple}{ regain} \textcolor{purple}{ Thr} \textcolor{purple}{ fourth}, \textcolor{purple}{ exclude} believe the \textcolor{purple}{ throw} \textcolor{purple}{ musician}ball is \textcolor{purple}{icester} \textcolor{purple}{ Lar} the \textcolor{purple}{ simplest} \textcolor{purple}{ outdoor} \textcolor{purple}{ activities}\\\textbf{\color{purple} Step: 999:}\\<s> As one of the founders of Bocce Ball in Holliston, I believe the four-ball is one of the most talented occasions\\
}
\end{AIbox}

\begin{AIbox}{}
\tt{\footnotesize  
\textbf{\color{purple} Step: 0:}\\<s>ION \textcolor{purple}{ wrote} \textcolor{purple}{ lasted} \textcolor{purple}{onial} \textcolor{purple}{thinking} \textcolor{purple}{ Pat} \textcolor{purple}{ric} \textcolor{purple}{kee} \textcolor{purple}{ly} \textcolor{purple}{ holog} \textcolor{purple}{ assures} \textcolor{purple}{ Bye} \textcolor{purple}{ rejo} \textcolor{purple}{ices} \textcolor{purple}{ec} \textcolor{purple}{onds} \textcolor{purple}{ dances} \textcolor{purple}{umi} \textcolor{purple}{GROUND} \textcolor{purple}{oubtedly} \textcolor{purple}{ oz} \textcolor{purple}{ handcuffed} \textcolor{purple}{ stamp} \textcolor{purple}{ateful} \textcolor{purple}{RG} \textcolor{purple}{ PlayStation} \textcolor{purple}{ mature}\\\textbf{\color{purple} Step: 250:}\\<s>ION \textcolor{purple}{istas} \textcolor{purple}{che} \textcolor{purple}{ acknowledged} \textcolor{purple}{ unto} \textcolor{purple}{ Developers} \textcolor{purple}{Fs} \textcolor{purple}{74} \textcolor{purple}{ Neigh} \textcolor{purple}{ rabbit} \textcolor{purple}{ chocolate} \textcolor{purple}{709} \textcolor{purple}{ …} \textcolor{purple}{ noise} \textcolor{purple}{ apply} \textcolor{purple}{ False} \textcolor{purple}{ sideways} \textcolor{purple}{ donors} \textcolor{purple}{ancy} \textcolor{purple}{ minimize} \textcolor{purple}{ offices} \textcolor{purple}{91} \textcolor{purple}{ update} \textcolor{purple}{ spider} \textcolor{purple}{ woods} \textcolor{purple}{ continually} \textcolor{purple}{olicy}\\\textbf{\color{purple} Step: 500:}\\<s>ION \textcolor{purple}{bb} \textcolor{purple}{ Charisma} \textcolor{purple}{ Depending} \textcolor{purple}{ Navajo} \textcolor{purple}{ UTF} \textcolor{purple}{ identified} \textcolor{purple}{ Him} \textcolor{purple}{ hazardous} \textcolor{purple}{ Gone} \textcolor{purple}{Denver} \textcolor{purple}{693} \textcolor{purple}{ clerk} \textcolor{purple}{2008} \textcolor{purple}{ overpowered} \textcolor{purple}{ warmed} \textcolor{purple}{ DL} \textcolor{purple}{ granted} \textcolor{purple}{yer} \textcolor{purple}{/-} \textcolor{purple}{ Rub} \textcolor{purple}{ ends} \textcolor{purple}{ believing} \textcolor{purple}{ brill} \textcolor{purple}{ Range} \textcolor{purple}{ nexus} \textcolor{purple}{ LSU}\\\textbf{\color{purple} Step: 750:}\\<s>ION COR \textcolor{purple}{ privileges}. \textcolor{purple}{112} \textcolor{purple}{ Hert} subsidiary \textcolor{purple}{ Diego} \textcolor{purple}{ HAM} \textcolor{purple}{RR} \textcolor{purple}{apego} \textcolor{purple}{SON}, DOLPHIN, \textcolor{purple}{ OL} \textcolor{purple}{IM} \textcolor{purple}{ERS} \textcolor{purple}{IB} \textcolor{purple}{AL} \textcolor{purple}{bsp} \textcolor{purple}{ M} \textcolor{purple}{IND} \textcolor{purple}{ICA}\\\textbf{\color{purple} Step: 999:}\\<s>ION CORP. is a subsidiary of DOLPHIN, DOLPHIN, FOLLOWARDILLARD, COROPD\\
}
\end{AIbox}

\subsection{Plaid}

\begin{AIbox}{}
\tt{\footnotesize \textbf{\color{purple} Step: 500:}\\ \textcolor{purple}{ation} \textcolor{purple}{.} \textcolor{purple}{ TM} \textcolor{purple}{ is} \textcolor{purple}{ used} \textcolor{purple}{ Weap} \textcolor{purple}{ improve} \textcolor{purple}{ with} \textcolor{purple}{ psych} \textcolor{purple}{al} \textcolor{purple}{ } \textcolor{purple}{ per} \textcolor{purple}{ro} \textcolor{purple}{ic}, councill \textcolor{purple}{ stress} \textcolor{purple}{ symptoms} \textcolor{purple}{ as} \textcolor{purple}{ well} \textcolor{purple}{ } \textcolor{purple}{ poster} \textcolor{purple}{ive} \textcolor{purple}{ disorders} \textcolor{purple}{.} \textcolor{purple}{ On} \textcolor{purple}{ent} \textcolor{purple}{il}\\\textbf{\color{purple} Step: 650:}\\ relaxation. \textcolor{purple}{ PT} is used to improve \textcolor{purple}{ skin} \textcolor{purple}{ ac} \textcolor{purple}{al}, \textcolor{purple}{ period} \textcolor{purple}{orage} \textcolor{purple}{,} \textcolor{purple}{ councill} \textcolor{purple}{ other} \textcolor{purple}{ areas} \textcolor{purple}{ and} \textcolor{purple}{ also} \textcolor{purple}{ potentially} \textcolor{purple}{ new} \textcolor{purple}{ scal} \textcolor{purple}{p} \textcolor{purple}{ growth} \textcolor{purple}{.} \textcolor{purple}{ One} \textcolor{purple}{ of}\\\textbf{\color{purple} Step: 700:}\\ \textcolor{purple}{ relax} \textcolor{purple}{ation} \textcolor{purple}{.} \textcolor{purple}{ ST} \textcolor{purple}{ is} \textcolor{purple}{ used} \textcolor{purple}{ to} \textcolor{purple}{ improve} \textcolor{purple}{ healthy} \textcolor{purple}{ post} \textcolor{purple}{ural} \textcolor{purple}{,} \textcolor{purple}{ pre} \textcolor{purple}{ens} \textcolor{purple}{inal} \textcolor{purple}{,} \textcolor{purple}{ and} \textcolor{purple}{ muscular} \textcolor{purple}{ muscles} \textcolor{purple}{ and} \textcolor{purple}{ help} \textcolor{purple}{ improved} \textcolor{purple}{ overall} \textcolor{purple}{ scal} \textcolor{purple}{vic} \textcolor{purple}{ function}. One\\\textbf{\color{purple} Step: 750:}\\ \textcolor{purple}{ation} \textcolor{purple}{.} \textcolor{purple}{ ST} \textcolor{purple}{ is} \textcolor{purple}{ designed} \textcolor{purple}{ to} \textcolor{purple}{ increase} \textcolor{purple}{ muscle} \textcolor{purple}{ flex} \textcolor{purple}{ility} \textcolor{purple}{,} \textcolor{purple}{ st} \textcolor{purple}{am} \textcolor{purple}{ina} \textcolor{purple}{,} \textcolor{purple}{ and} \textcolor{purple}{ core} \textcolor{purple}{ strength} \textcolor{purple}{ and} \textcolor{purple}{ also} \textcolor{purple}{ promote} \textcolor{purple}{ overall} \textcolor{purple}{ a} \textcolor{purple}{erv} \textcolor{purple}{ic} \textcolor{purple}{ growth}. One\\\textbf{\color{purple} Step: 800:}\\ relaxation. ST is designed to increase mental acuity, stamina, and physical strength and help reduce \textcolor{purple}{ major} depressive symptoms. \textcolor{purple}{ One}\\\textbf{\color{purple} Step: 850:}\\ relaxation. ST is designed to \textcolor{purple}{ increase} mental acuity, stamina, and physical strength and help \textcolor{purple}{ reduce} \textcolor{purple}{ past} depressive symptoms. Some\\\textbf{\color{purple} Step: 900:}\\ relaxation. ST is designed to \textcolor{purple}{ improve} mental acuity, stamina, and physical strength and help alleviate major depressive symptoms. Some\\\textbf{\color{purple} Step: 950:}\\ relaxation. ST is designed to increase mental acuity, stamina, and \textcolor{purple}{ physical} \textcolor{purple}{ strength} and help alleviate major depressive symptoms. Some\\\textbf{\color{purple} Step: 999:}\\ relaxation. ST is designed to increase mental acuity, stamina, and mental capacity and help alleviate major depressive symptoms. Some\\
}

\end{AIbox}

\begin{AIbox}{}
\tt{\footnotesize \textbf{\color{purple} Step: 500:}\\ \textcolor{purple}{.} \textcolor{purple}{ With} \textcolor{purple}{ conclud} \textcolor{purple}{ p} \textcolor{purple}{ors} \textcolor{purple}{ still} \textcolor{purple}{ made} \textcolor{purple}{ in} \textcolor{purple}{ other} \textcolor{purple}{ times} \textcolor{purple}{,} \textcolor{purple}{ this} \textcolor{purple}{ number} \textcolor{purple}{ntil} \textcolor{purple}{ war} \textcolor{purple}{ will} \textcolor{purple}{ be} \textcolor{purple}{ more} \textcolor{purple}{ enough} \textcolor{purple}{ by} \textcolor{purple}{ pin} \textcolor{purple}{ and} \textcolor{purple}{ others} \textcolor{purple}{.} \textcolor{purple}{ But} \textcolor{purple}{ conclud} \textcolor{purple}{ game} \textcolor{purple}{ itself}\\\textbf{\color{purple} Step: 650:}\\ \textcolor{purple}{ conclud} \textcolor{purple}{ padd} \textcolor{purple}{ters} only \textcolor{purple}{ move} in \textcolor{purple}{ high} \textcolor{purple}{ speed}, no \textcolor{purple}{ number} \textcolor{purple}{ntil} \textcolor{purple}{ players} \textcolor{purple}{ can} be more fun to \textcolor{purple}{ car} \textcolor{purple}{osate} than \textcolor{purple}{ others}. \textcolor{purple}{ In} the game you can\\\textbf{\color{purple} Step: 700:}\\ \textcolor{purple}{ a} \textcolor{purple}{ ty} \textcolor{purple}{ter} \textcolor{purple}{ only} \textcolor{purple}{ living} in \textcolor{purple}{ small} \textcolor{purple}{ times}, no form of \textcolor{purple}{ game} \textcolor{purple}{ is} be more \textcolor{purple}{ fun} \textcolor{purple}{ to} \textcolor{purple}{ beh} \textcolor{purple}{o} than \textcolor{purple}{ football}. \textcolor{purple}{ Over} the \textcolor{purple}{ game} you can\\\textbf{\color{purple} Step: 750:}\\ \textcolor{purple}{ conclud} \textcolor{purple}{ ump} \textcolor{purple}{ires} \textcolor{purple}{ still} \textcolor{purple}{ played} \textcolor{purple}{ in} \textcolor{purple}{ human} sports, no \textcolor{purple}{ form} of \textcolor{purple}{ sport} \textcolor{purple}{ can} be more perfect for \textcolor{purple}{ bats}ankind than \textcolor{purple}{ cricket}. In the \textcolor{purple}{ years} \textcolor{purple}{ you} \textcolor{purple}{ can}\\\textbf{\color{purple} Step: 800:}\\ a \textcolor{purple}{ rept} \textcolor{purple}{ile} \textcolor{purple}{ now} \textcolor{purple}{ focused} \textcolor{purple}{ for} traditional sports, no type of game could be more perfect for butter \textcolor{purple}{ankind} than football. \textcolor{purple}{ In} the \textcolor{purple}{ world} we \textcolor{purple}{ are}\\\textbf{\color{purple} Step: 850:}\\ a \textcolor{purple}{ land} \textcolor{purple}{ator} already \textcolor{purple}{ adept} \textcolor{purple}{ in} \textcolor{purple}{ traditional} sports, no type of game \textcolor{purple}{ could} be more perfect for \textcolor{purple}{ butter} \textcolor{purple}{ria} than football. \textcolor{purple}{ Around} the country we have\\\textbf{\color{purple} Step: 900:}\\ a \textcolor{purple}{ gard} \textcolor{purple}{ener} \textcolor{purple}{ already} \textcolor{purple}{ dependent} \textcolor{purple}{ with} \textcolor{purple}{ modern} sports, no \textcolor{purple}{ type} of game \textcolor{purple}{ can} be more perfect for beekeeping than football. Across the \textcolor{purple}{ country} we have\\\textbf{\color{purple} Step: 950:}\\ a beekeeper also interested in environmental sports, no \textcolor{purple}{ style} of game could be more perfect for beekeeping than \textcolor{purple}{ football}. \textcolor{purple}{ Across} the state we have\\\textbf{\color{purple} Step: 999:}\\ a beekeeper also interested in environmental sports, no type of game could be more perfect for beekeeping than baseball. Around the state we have\\
}

\end{AIbox}

\begin{table*}[ht!]
\begin{center}
\begin{small}

\begin{tabular}{r|c|c|ccccc}
\toprule
Task   & TW                    & $t_{\text{max}}$      & AR-NLL        & dist-1                      & MAUVE & self-BLEU     & zipf \\
\midrule
Data   & -                     & -                     & 3.29          & N/A                         & N/A   & 0.09          & 0.86 \\
\midrule
\multicolumn{8}{c}{Unconditional}                                                                                                   \\
\midrule

Span   &                       &                       & 3.89          & \textbf{0.54}               & N/A   & 0.27          & 1.01     \\
MLM    &                       &                       & 3.83          & 0.50                        & N/A   & 0.34          & 1.19     \\
Prefix & \multirow{-3}{*}{No}  &                       & 4.06          & 0.53                        & N/A   & \textbf{0.24} & 0.99     \\

Span   &                       &                       & 3.92          & 0.52                        & N/A   & \textbf{0.24} & 1.00  \\
MLM    &                       &                       & \textbf{3.72} & 0.50                        & N/A   & 0.34          & 1.28 \\

Prefix & \multirow{-3}{*}{Yes} & \multirow{-6}{*}{10}  & 3.82          & 0.53                        & N/A   & 0.27          & 1.13 \\
\midrule
\multicolumn{8}{c}{Prefix-32}                                                                                                                    \\
\midrule
Span   &                       &                       & 3.77          & \textbf{0.57}               & 0.91          & \textbf{0.14} & 0.88          \\
MLM    &                       &                       & 3.70          & 0.55                        & 0.86          & 0.16          & 0.90          \\
Prefix & \multirow{-3}{*}{No}  &                       & 3.78          & \textbf{0.57}               & 0.89          & 0.15          & 0.88          \\
Span   &                       &                       & 3.77          & 0.56                        & \textbf{0.92} & \textbf{0.14} & \textbf{0.87} \\
MLM    &                       &                       & \textbf{3.65} & 0.54                        & 0.86          & 0.15          & 0.91          \\
Prefix & \multirow{-3}{*}{Yes} & \multirow{-6}{*}{10}  & 3.75          & \textbf{0.57}               & 0.91          & 0.15          & 0.89          \\
\midrule
\multicolumn{8}{c}{Enclosed-32}                                                                                                                      \\
\midrule
Span   &                       &                       & 3.82          & 0.57                        & \textbf{0.92} & 0.16          & 0.89          \\
MLM    &                       &                       & 3.74          & 0.55                        & 0.91          & 0.17          & 0.90          \\
Prefix & \multirow{-3}{*}{No}  &                       & 3.89          & 0.57                        & 0.91          & 0.16          & 0.88          \\
Span   &                       &                       & 3.84          & 0.57                        & 0.91          & \textbf{0.15} & \textbf{0.87} \\
MLM    &                       &                       & \textbf{3.69} & 0.54                        & 0.90          & 0.17          & 0.91          \\
Prefix & \multirow{-3}{*}{Yes} & \multirow{-6}{*}{10}  & 3.86          & \textbf{0.58}               & 0.91          & 0.16          & 0.90          \\
\bottomrule
\end{tabular}
        \caption{Evaluation of DDLM with different masking strategies, $t_{\text{max}} = 10$, and with/without time warping for Unconditional, Prefix-32, and Enclosed-32 generation settings. We bolded the best metric values across other runs. See Section \ref{section:ddlm-training} for more details. See Appendix Tables \ref{tab:ablation-prefix}, \ref{tab:ablation-unconditional}, \ref{tab:ablation-enclosed} for the full list of results with a wider range of $t_{\text{max}}$ values.}
                    \label{tab:ablation}
\end{small}
\end{center}
\end{table*}

\begin{table*}[ht!]
\begin{center}
\begin{small}

\begin{tabular}{r|c|c|ccccc}
\toprule
\multicolumn{8}{c}{Unconditional}                                                                                                   \\
\toprule
Task   & TW                    & $t_{\text{max}}$      & AR-NLL        & dist-1                      & MAUVE & self-BLEU     & zipf \\
\toprule
Data   & -                     & -                     & 3.29          & N/A                         & N/A   & 0.09          & 0.86 \\
\midrule

Span   &                       &                       & 3.89          & \textbf{0.54}               & N/A   & 0.27          & 1.01     \\
MLM    &                       &                       & 3.83          & 0.50                        & N/A   & 0.34          & 1.19     \\
Prefix & \multirow{-3}{*}{No}  &                       & 4.06          & 0.53                        & N/A   & \textbf{0.24} & 0.99     \\

Span   &                       &                       & 3.92          & 0.52                        & N/A   & \textbf{0.24} & 1.00  \\
MLM    &                       &                       & \textbf{3.72} & 0.50                        & N/A   & 0.34          & 1.28 \\

Prefix & \multirow{-3}{*}{Yes} & \multirow{-6}{*}{10}  & 3.82          & 0.53                        & N/A   & 0.27          & 1.13 \\
\midrule
Span   &                       &                       & 2.13          & {\color[HTML]{CB0000} 0.20} & N/A   & 0.84          & 1.81 \\
MLM    &                       &                       & 2.96          & {\color[HTML]{CB0000} 0.19} & N/A   & 0.81          & 1.70 \\
Prefix & \multirow{-3}{*}{No}  &                       & 2.11          & {\color[HTML]{CB0000} 0.19} & N/A   & 0.89          & 1.98 \\
Span   &                       &                       & 2.19             & {\color[HTML]{CB0000} 0.24}     & N/A   &   0.80            &  1.78    \\
MLM    &                       &                       & 3.04          & {\color[HTML]{CB0000} 0.04} & N/A   & 0.96          & 2.33 \\
Prefix & \multirow{-3}{*}{Yes} & \multirow{-6}{*}{50}  & 2.11          & {\color[HTML]{CB0000} 0.22} & N/A   & 0.77          & 1.76 \\
\midrule
Span   &                       &                       & 2.97              & {\color[HTML]{CB0000} 0.04}     & N/A   &  0.99             &  3.50    \\
MLM    &                       &                       & 3.00          & {\color[HTML]{CB0000} 0.04} & N/A   & 0.99          & 3.69 \\
Prefix & \multirow{-3}{*}{No}  &                       & 1.42          & {\color[HTML]{CB0000} 0.01} & N/A   & 0.99          & 3.49 \\
Span   &                       &                       & 1.73          & {\color[HTML]{CB0000} 0.14} & N/A   & 0.95          & 2.59 \\
MLM    &                       &                       & 1.10          & {\color[HTML]{CB0000} 0.01} & N/A   & 0.99          & 5.10 \\
Prefix & \multirow{-3}{*}{Yes} & \multirow{-6}{*}{300} & 2.14          & {\color[HTML]{CB0000} 0.07} & N/A   & 0.98          & 3.01 \\
\bottomrule
\end{tabular}
        \caption{Evaluation of DDLM with different masking strategies, $t_{\text{max}}$ values, and with/without time warping for Unconditional generation setting. The metrics with values $< 0.5$ (indicating highly repetitive samples) are displayed in colored font. We bolded the best metric values across other runs. See Section \ref{section:ddlm-training} for more details.}
                    \label{tab:ablation-unconditional}
\end{small}
\end{center}
\end{table*}

\begin{table*}[ht!]
\begin{center}
\begin{small}

\begin{tabular}{r|c|c|ccccc}
\toprule
\multicolumn{8}{c}{Prefix-32}                                                                                                                    \\
\toprule
Task   & TW                    & $t_{\text{max}}$      & AR-NLL        & dist-1                      & MAUVE         & self-BLEU     & zipf          \\
\toprule
Data   & -                     & -                     & 3.29          & N/A                         & N/A           & 0.09          & 0.86          \\
\midrule
Span   &                       &                       & 3.77          & \textbf{0.57}               & 0.91          & \textbf{0.14} & 0.88          \\
MLM    &                       &                       & 3.70          & 0.55                        & 0.86          & 0.16          & 0.90          \\
Prefix & \multirow{-3}{*}{No}  &                       & 3.78          & \textbf{0.57}               & 0.89          & 0.15          & 0.88          \\
Span   &                       &                       & 3.77          & 0.56                        & \textbf{0.92} & \textbf{0.14} & \textbf{0.87} \\
MLM    &                       &                       & \textbf{3.65} & 0.54                        & 0.86          & 0.15          & 0.91          \\
Prefix & \multirow{-3}{*}{Yes} & \multirow{-6}{*}{10}  & 3.75          & \textbf{0.57}               & 0.91          & 0.15          & 0.89          \\
\midrule
Span   &                       &                       & 3.31          & {\color[HTML]{CB0000} 0.27} & 0.67          & 0.24          & 0.90          \\
MLM    &                       &                       & 3.27          & {\color[HTML]{CB0000} 0.27} & 0.79          & 0.15          & 0.85          \\
Prefix & \multirow{-3}{*}{No}  &                       & 3.24          & {\color[HTML]{CB0000} 0.26} & 0.63          & 0.27          & 0.92          \\
Span   &                       &                       & 3.07              & {\color[HTML]{CB0000} 0.25}     &    0.70           &     0.21          &    0.89           \\
MLM    &                       &                       & 3.06          & {\color[HTML]{CB0000} 0.27} & 0.76          & 0.18          & 0.87          \\
Prefix & \multirow{-3}{*}{Yes} & \multirow{-6}{*}{50}  & 3.11          & {\color[HTML]{CB0000} 0.26} & 0.78          & 0.19          & 0.89          \\
\midrule
Span   &                       &                       & 3.59              & {\color[HTML]{CB0000} 0.12}     &        0.05       &     0.26          &   1.01            \\
MLM    &                       &                       & 3.96          & {\color[HTML]{CB0000} 0.14} & 0.07          & 0.20          & 0.98          \\
Prefix & \multirow{-3}{*}{No}  &                       & 3.28          & {\color[HTML]{CB0000} 0.11} & 0.05          & 0.38          & 1.00          \\
Span   &                       &                       & 3.06          & {\color[HTML]{CB0000} 0.14} & 0.28          & 0.27          & 0.97          \\
MLM    &                       &                       & 3.37          & {\color[HTML]{CB0000} 0.15} & 0.15          & 0.27          & 0.95          \\
Prefix & \multirow{-3}{*}{Yes} & \multirow{-6}{*}{300} & 3.11          & {\color[HTML]{CB0000} 0.13} & 0.22          & 0.33          & 0.99          \\
\bottomrule
\end{tabular}
        \caption{Evaluation of DDLM with different masking strategies, $t_{\text{max}}$ values, and with/without time warping for Prefix-32 generation setting. The metrics with values $< 0.5$ (indicating highly repetitive samples) are displayed in colored font. We bolded the best metric values across other runs. See Section \ref{section:ddlm-training} for more details.}
                    \label{tab:ablation-prefix}
\end{small}
\end{center}
\end{table*}

\begin{table*}[ht!]
\begin{center}
\begin{small}

\begin{tabular}{r|c|c|ccccc}
\toprule
\multicolumn{8}{c}{Enclosed-32}                                                                                                                      \\
\toprule
Task   & TW                    & $t_{\text{max}}$      & AR-NLL        & dist-1                      & MAUVE         & self-BLEU     & zipf          \\
\midrule
Data   & -                     & -                     & 3.29          & N/A                         & N/A           & 0.09          & 0.86          \\
\midrule
Span   &                       &                       & 3.82          & 0.57                        & \textbf{0.92} & 0.16          & 0.89          \\
MLM    &                       &                       & 3.74          & 0.55                        & 0.91          & 0.17          & 0.90          \\
Prefix & \multirow{-3}{*}{No}  &                       & 3.89          & 0.57                        & 0.91          & 0.16          & 0.88          \\
Span   &                       &                       & 3.84          & 0.57                        & 0.91          & \textbf{0.15} & \textbf{0.87} \\
MLM    &                       &                       & \textbf{3.69} & 0.54                        & 0.90          & 0.17          & 0.91          \\
Prefix & \multirow{-3}{*}{Yes} & \multirow{-6}{*}{10}  & 3.86          & \textbf{0.58}               & 0.91          & 0.16          & 0.90          \\
\midrule
Span   &                       &                       & 3.35          & {\color[HTML]{CB0000} 0.29} & 0.90          & 0.24          & 0.90          \\
MLM    &                       &                       & 3.34          & {\color[HTML]{CB0000} 0.29} & 0.90          & 0.16          & 0.86          \\
Prefix & \multirow{-3}{*}{No}  &                       & 3.41          & {\color[HTML]{CB0000} 0.27} & 0.90          & 0.30          & 0.94          \\
Span   &                       &                       & 3.14              & {\color[HTML]{CB0000} 0.27}     &         0.91      &    0.23           &   0.90            \\
MLM    &                       &                       & 3.12          & {\color[HTML]{CB0000} 0.29} & 0.90          & 0.19          & 0.87          \\
Prefix & \multirow{-3}{*}{Yes} & \multirow{-6}{*}{50}  & 3.26          & {\color[HTML]{CB0000} 0.27} & 0.90          & 0.21          & 0.89          \\
\midrule
Span   &                       &                       & 3.66              & {\color[HTML]{CB0000} 0.15}     &     0.91         &    0.33           &    1.01           \\
MLM    &                       &                       & 3.93          & {\color[HTML]{CB0000} 0.17} & 0.89          & 0.21          & 0.95          \\
Prefix & \multirow{-3}{*}{No}  &                       & 3.40          & {\color[HTML]{CB0000} 0.12} & 0.90          & 0.40          & 1.02          \\
Span   &                       &                       & 3.21          & {\color[HTML]{CB0000} 0.17} & 0.90          & 0.24          & 0.94          \\
MLM    &                       &                       & 3.38          & {\color[HTML]{CB0000} 0.18} & 0.90          & 0.24          & 0.91          \\
Prefix & \multirow{-3}{*}{Yes} & \multirow{-6}{*}{300} & 3.25          & {\color[HTML]{CB0000} 0.14} & 0.90          & 0.34          & 1.00         \\
\bottomrule
\end{tabular}
        \caption{Evaluation of DDLM with different masking strategies, $t_{\text{max}}$ values, and with/without time warping for Enclosed-32 generation setting. The metrics with values $< 0.5$ (indicating highly repetitive samples) are displayed in colored font. We bolded the best metric values across other runs. See Section \ref{section:ddlm-training} for more details.}
                    \label{tab:ablation-enclosed}

\end{small}
\end{center}
\end{table*}

\end{document}

%% file: math_commands.tex
\usepackage{amsmath,amsfonts,bm}

\def\eqref#1{equation~\ref{#1}}

\def\1{\bm{1}}

\def\vx{{\bm{x}}}

\def\mS{{\bm{S}}}

\def\mV{{\bm{V}}}

\def\mX{{\bm{X}}}

\DeclareMathAlphabet{\mathsfit}{\encodingdefault}{\sfdefault}{m}{sl}
\SetMathAlphabet{\mathsfit}{bold}{\encodingdefault}{\sfdefault}{bx}{n}

\newcommand{\R}{\mathbb{R}}

